\documentclass[letterpaper,  paper,11pt]{AAS}	

\usepackage{bm}
\usepackage{amsmath}
\usepackage{subfigure}
\usepackage[colorlinks=true, pdfstartview=FitV, linkcolor=black, citecolor= black, urlcolor= black]{hyperref}
\usepackage{overcite}
\usepackage{footnpag}			      	
\usepackage{wrapfig}
\usepackage{float}
\makeatletter
\newcommand{\figcaption}[1]{\def\@captype{figure}\caption{#1}}
\newcommand{\tblcaption}[1]{\def\@captype{table}\caption{#1}}
\makeatother

\PaperNumber{25-372}

\begin{document}

\title{A Prefixed Patch Time Series Transformer for Two-Point Boundary Value Problems in Three-Body Problems}

\author{Akira Hatakeyama\thanks{Ph.D. student, Institute of Space and Astronautical Science, Japan Aerospace Exploration Agency, The Graduate University for Advanced Studies, SOKENDAI., 240-0193.},  
Shota Ito\thanks{Ph.D. student, Department of Aerospace Engineering, Tokyo Metropolitan University, 192-0397.},
Toshihiko Yanase\thanks{Engineer, AI Computing Division, Preferred Networks, Inc., 100-0004.},
\ and Naoya Ozaki\thanks{Associate Professor, Department of Spacecraft Engineering, Institute of Space and Astronautical Science, Japan Aerospace Exploration Agency, 252-5210.}
}

\maketitle{}

\begin{abstract}
Two-point boundary value problems for cislunar trajectories present significant challenges in circler restricted three body problem, making traditional analytical methods like Lambert's problem inapplicable. This study proposes a novel approach using a prefixed patch time series Transformer model that automates the solution of two-point boundary value problems from lunar flyby to arbitrary terminal conditions. Using prefix tokens of terminal conditions in our deep generative model enables solving boundary value problems in three-body dynamics. The training dataset consists of trajectories obtained through forward propagation rather than solving boundary value problems directly. The model demonstrates potential practical utility for preliminary trajectory design in cislunar mission scenarios.
\end{abstract}

\section{Introduction}
The analysis of three-body dynamics is fundamental to advancing the capabilities of cislunar trajectory optimization and mission architecture design. In contrast to two-body systems, which admit analytical solutions, the three-body problem is characterized by nonintegrable dynamics that preclude closed-form solutions. Although significant theoretical advances have been made in solving two-point boundary value problems—particularly their reduction to Lambert's problem within two-body frameworks—the extension of these methodologies to three-body systems presents substantial computational challenges that have yet to be effectively resolved.\cite{ken2020}

Over recent decades, the field of astrodynamics has shown emerging progress through the integration of machine learning methodologies into astrodynamics. Studies have shown the effectiveness of using generative models in unknown tasks and global trajectory design problems by controlling low-thrust propulsion through meta-reinforcement learning\cite{Fe2024}. \cite{sullivan2020using} \cite{Zavoli2021reinforcement}. In addition, research has demonstrated the utility of deep generative models for trajectory design using a Conditional Variational AutoEncoder\cite{li2023amortized}. Recent developments include the use of deep neural networks for the autonomous mission design, and the application of supervised learning to trajectory optimization \cite{Izzo2021}. Further advances have been made using Transformers to predict control variables and equations of motion for trajectory propagation\cite{Guffanti2024transformers} \cite{Presser2024}. Despite these advances in deep generative model approaches \cite{briden2025diffusion}, there is no method to generate initial guess trajectories for trajectory optimization problems in the three-body problem.

In the field of machine learning, time series analysis and forecasting methods have evolved beyond traditional statistical approaches. With the success of Large Language Models (LLMs) like ChatGPT transforming sequential data processing, time series forecasting has also seen a shift from classical methods like autoregressive models \cite{box2015time} and moving average models \cite{McKenzie1984} to deep generative model approaches. The introduction of the Transformer architecture by Vaswani et al.\cite{NIPS2017_3f5ee243}, which revolutionized sequential data processing with its attention mechanism, has led to significant advances in various domains. Following this breakthrough, Transformer-based models have emerged as powerful tools for time series forecasting, demonstrating superior performance compared to traditional approaches. The development of novel architectures has produced models with remarkable results in benchmarking, such as \cite{oreshkin2019n} with N-BEATS, \cite{das2023decoder} with Google's TimesFM, and \cite{Nie2022} with Patch Time Series Transformer (PatchTST). In particular, PatchTST demonstrated remarkable performance on standard benchmarks with its simple yet effective approach of treating time series as patches to the Transformer model, influencing subsequent research such as TimesFM. These advances, as well as the large language models by \cite{kenton2019bert} and \cite{brown2020language}, have shown the potential to handle time series prediction and generation. While these methods show promise, their application to nonlinear dynamical systems remains an open challenge in spacecraft trajectory design.

This study extends the PatchTST architecture to solve two-point boundary value problems in orbital dynamics, specifically addressing the fundamental three-body problem. The key innovation is the introduction of prefix conditioning, where initial and terminal states are encoded as prefix tokens to generate trajectories connecting these boundary points. Through hyperparameter optimization and analysis of 100 generated trajectories, we statistically evaluated the model's performance by measuring position and velocity errors of the generated solutions in the circularly constrained three-body problem.

\section{Fundamentals}
\subsection{Dynamical System}

\begin{figure}
   \centering
   \includegraphics[width=0.75\linewidth]{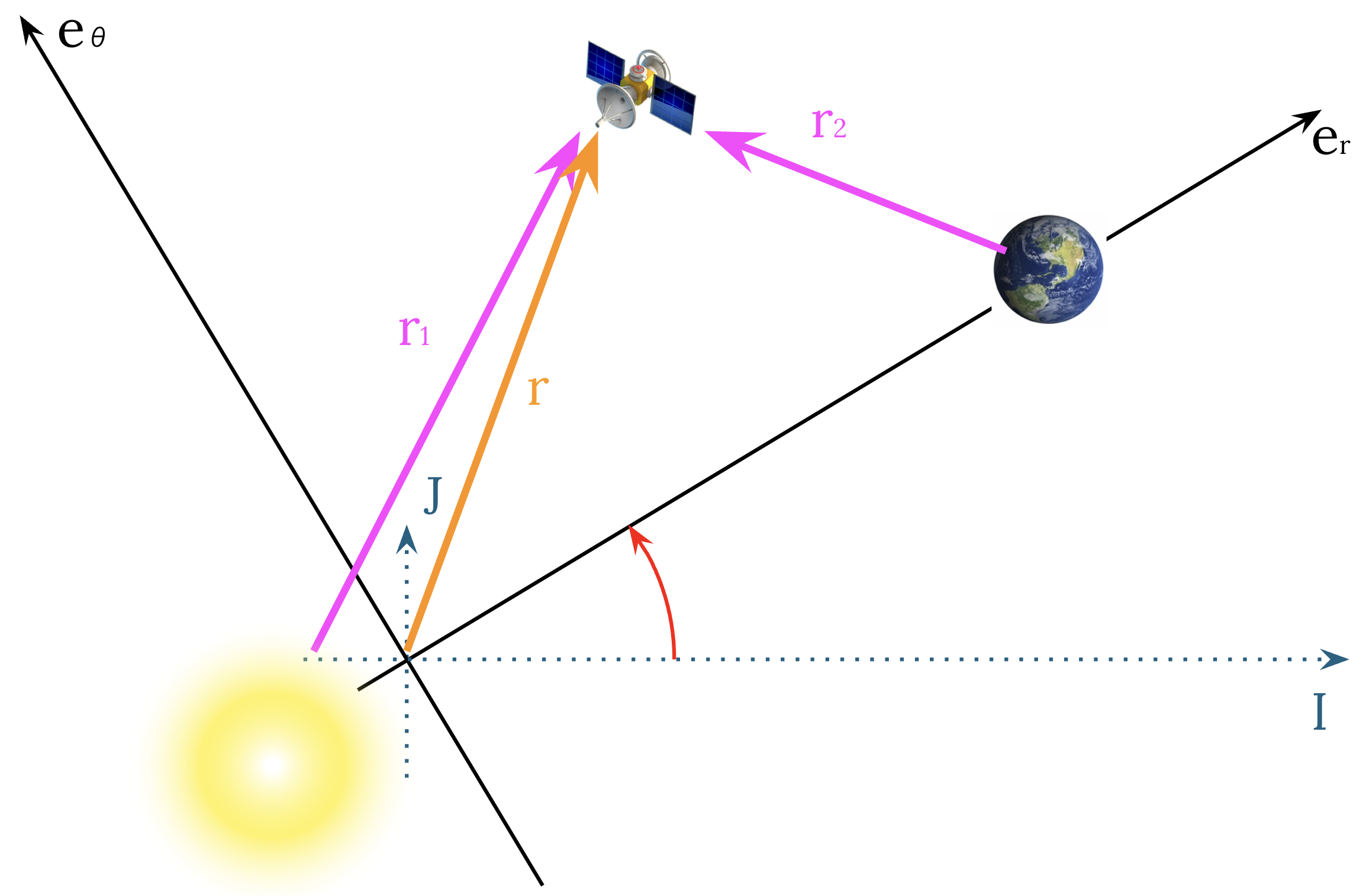}
   \caption{Sun-Earth-Spacecraft CR3BP}
   \label{fig:cr3bp}
\end{figure}

The dynamics of spacecraft in the cislunar region can be modeled using the Circular Restricted Three-Body Problem (CR3BP), as illustrated in Figure~\ref{fig:cr3bp}. In this study, we consider the Sun-Earth CR3BP, where two primary bodies (the Sun and Earth) move in circular orbits around their common center of mass. This model is particularly suitable for analyzing Earth-departure trajectories incorporating lunar flybys, such as those used in Artemis and CLPS ride-share opportunities, are suited for analysis using this model.

In the zero-sphere-of-influence approximation, the gravitational influence of the Moon is neglected during the analysis of the three-body dynamics. The mass of the spacecraft is assumed to be negligible compared to the mass of the primary bodies. The equations of motion in the non-dimensional, rotating coordinate system are given by the following equations:

\begin{equation}
\label{EoM}
\begin{aligned}
\begin{bmatrix}
\dot{x} \\
\dot{y} \\
\dot{z} \\
\dot{v}_x \\
\dot{v}_y \\
\dot{v}_z
\end{bmatrix}
&=
\begin{bmatrix}
v_x \\
v_y \\
v_z \\
2v_y + x - \frac{(1-\mu)(x+\mu)}{r_1^3} - \frac{\mu(x-1+\mu)}{r_2^3} \\
-2v_x + y - \frac{(1-\mu)y}{r_1^3} - \frac{\mu y}{r_2^3} \\
-\frac{(1-\mu)z}{r_1^3} - \frac{\mu z}{r_2^3}
\end{bmatrix} \\
r_1 &= \sqrt{(x+\mu)^2 + y^2 + z^2}, \quad r_2 = \sqrt{(x-1+\mu)^2 + y^2 + z^2}
\end{aligned}
\end{equation}

where $\mu$ is the mass parameter of the Sun-Earth system, $(x, y, z)$ is the position of the spacecraft in the rotating frame, and $(v_x, v_y, v_z)$ are the corresponding velocities. The distances $r_1$ and $r_2$ are the distances of the spacecraft to the Sun and Earth, respectively. In this study, we numerically integrate these ODEs to generate time series data of spacecraft trajectories.

\subsection{Transformer Architecture}
\begin{figure}[htb]
\centering
\includegraphics[width=0.8\linewidth]{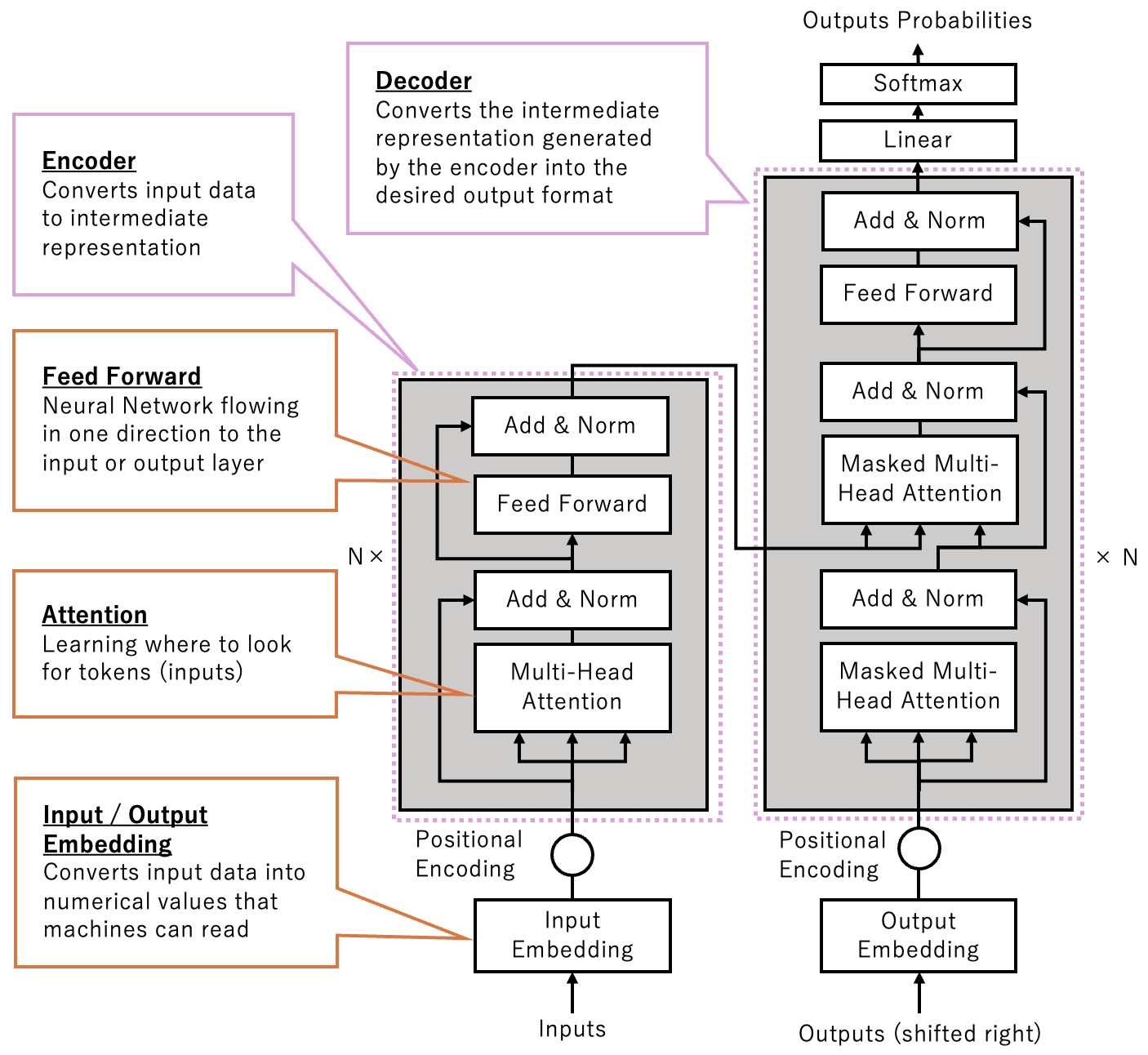}
\caption{Overview of Transformer Model}
\label{fig:transformer}
\end{figure}

Our generative model uses the Transformer architecture as its foundation, as shown in Figure~\ref{fig:transformer}~\cite{NIPS2017_3f5ee243}. The model consists of an encoder and a decoder component, where the encoder processes the input data into a latent space representation, and the decoder generates the output data from this latent space. The architecture allows us to perform a variety of challenging tasks such as translation, image generation, music generation, and text summarization. At the core of the model is the attention mechanism, which computes the relevance of different parts of the input sequence for each element of the output sequence.

\subsection{Patch Time Series Transformer}
The Patch Time Series Transformer (PatchTST) extends the standard Transformer architecture by introducing a patch-based processing approach for time series data \cite{Nie2022}. Instead of processing individual time steps, PatchTST segments the input time series into fixed-length patches that serve as token inputs to the Transformer.

\begin{figure}[htb]
    \centering
    \includegraphics[width=0.75\linewidth]{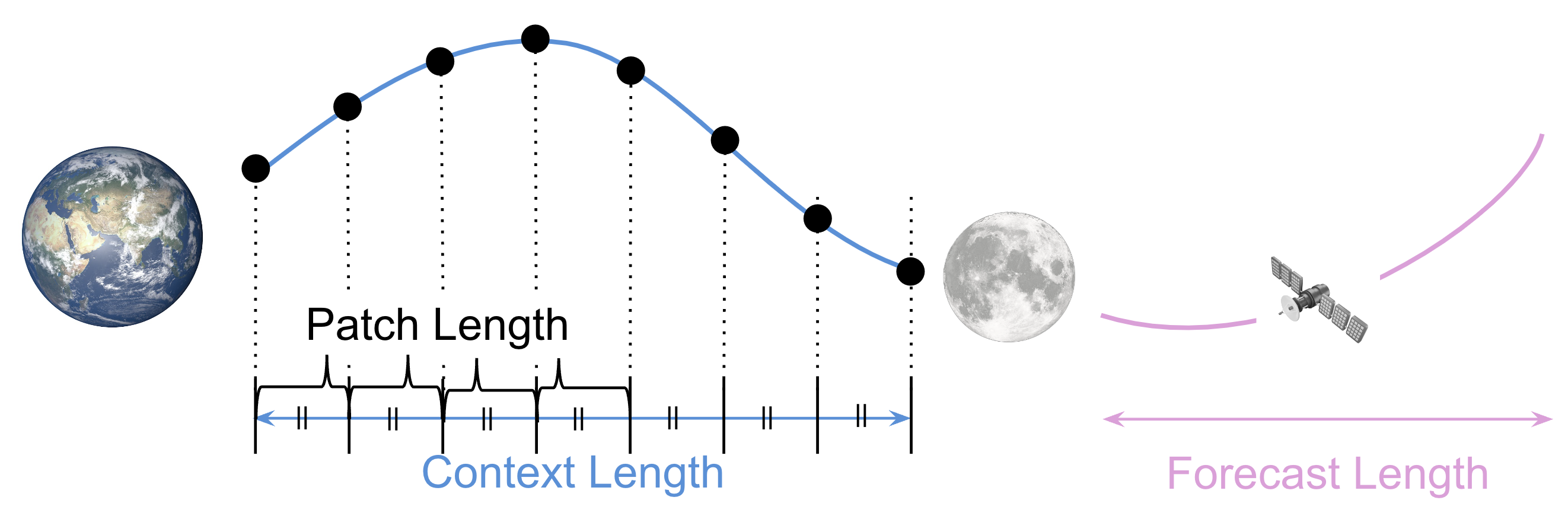}
    \caption{Illustration of key components in PatchTST}
    \label{fig:patch_concept}
\end{figure}

For orbital trajectory prediction, the architecture operates on three key parameters, also shown in Figure~\ref{fig:patch_concept}:
\begin{itemize}
    \item Context Length: The temporal span of historical trajectory data containing spacecraft state vectors (position and velocity)
    \item Patch Length: The number of time steps aggregated into each input token
    \item Forecast Length: The prediction horizon for future trajectory states
\end{itemize}

\begin{figure}[htb]
    \centering
    \includegraphics[width=\linewidth]{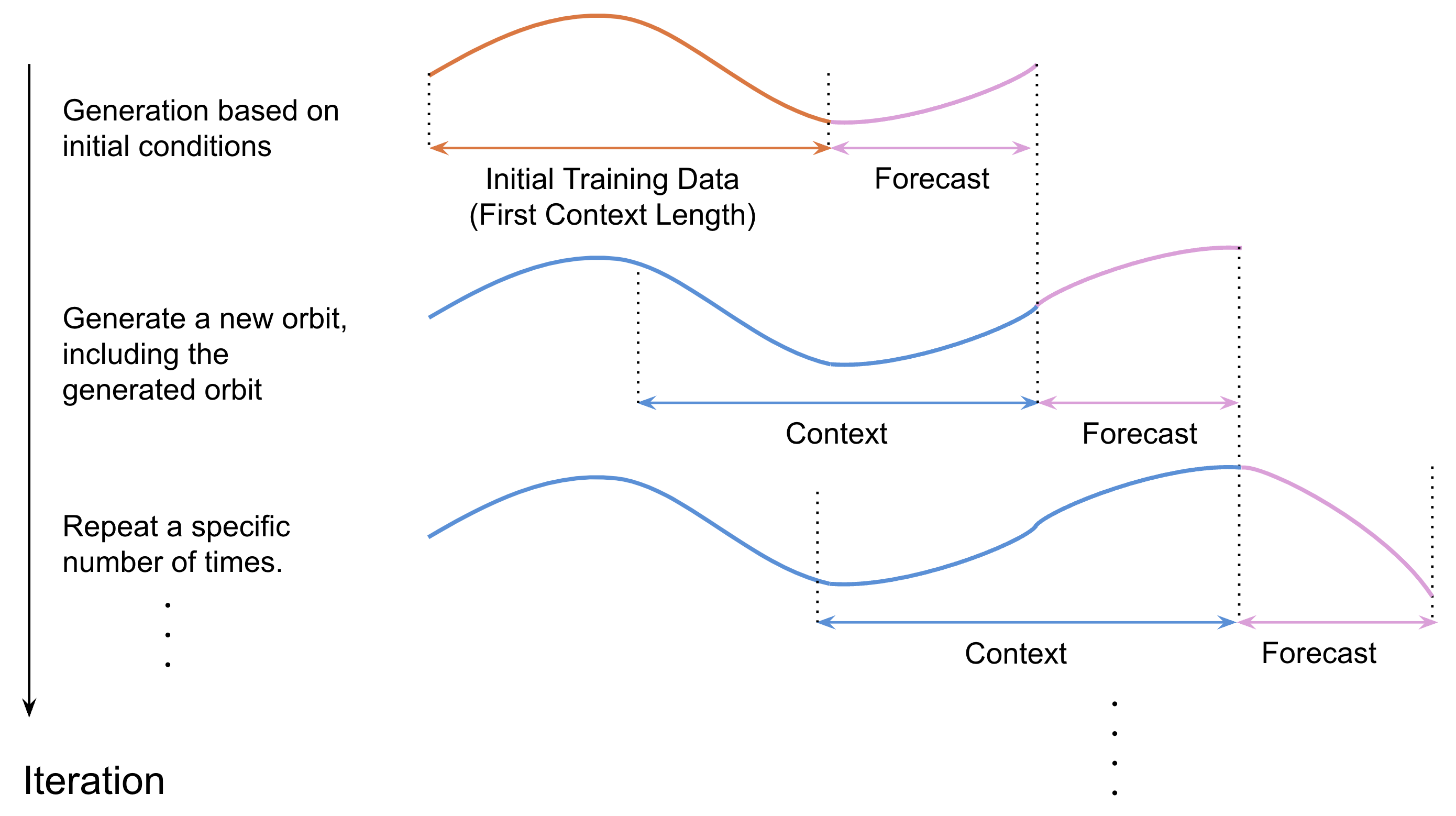}
    \caption{Iterative generation process in PatchTST}
    \label{fig:patch_iteration}
\end{figure}

The model employs an iterative generation process (Figure~\ref{fig:patch_iteration}) where predictions from earlier iterations become part of the context window for subsequent forecasts. This approach enables the model to generate extended trajectory predictions while maintaining numerical stability through the patch-based processing mechanism.

\section{Proposed method}
\subsection{Overview}
\begin{figure}[htb]
\centering
\includegraphics[width=1.0\linewidth]{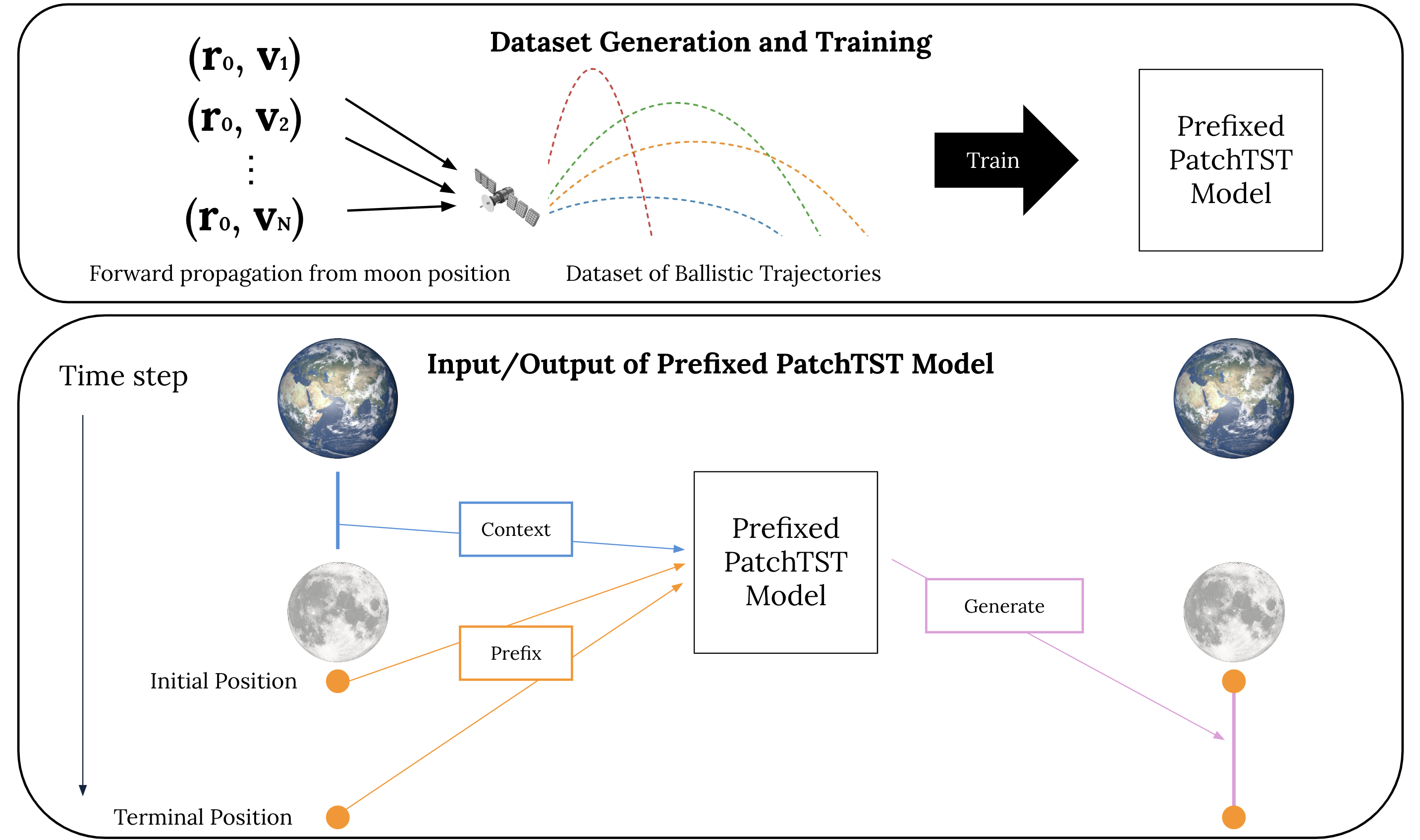}
\caption{Overview of the proposed method}
\label{fig:implementation}
\end{figure}

We generate the initial guess trajectories as a two-point boundary value problem, given the initial state and the spacecraft's desired terminal position, with gravity assist maneuvers around the Moon incorporated along the trajectory. Our goal is to determine the complete trajectory that satisfies boundary condition. Figure~\ref{fig:implementation} shows the general framework of our approach. The key innovation is the introduction of prefix tokens that allow the model to handle boundary conditions.

\subsection{Lunar Flyby Dataset Generation}

The trajectory dataset was generated using a six-dimensional state vector (position and velocity) as time-series data under the Zero-Sphere of Influence (ZeroSOI) patched conics assumption. The spacecraft's initial position was set to coincide with the Moon's position at flyby. For initial velocities, we parameterized the hyperbolic excess velocity ($V_{\infty}$) by systematically varying its direction while maintaining its magnitude, then adding it to the spacecraft's velocity.

To satisfy PatchTST model requirements, we first generated reference trajectories covering the required initial context length, representing viable Earth-to-Moon transfer orbits. These reference trajectories were computed through backward propagation from the lunar encounter point, using specified incoming $V_{\infty}$ parameters to ensure Earth-approaching trajectories. This reference set remained consistent across all cases.

The dataset was expanded by generating multiple trajectory variants through forward propagation from the lunar encounter, with outgoing $V_{\infty}$ parameters constrained by the relationship with incoming $V_{\infty}$ as detailed in Appendix A. Random shuffling was applied to the final dataset to mitigate potential training biases.

\subsection{Prefixed PatchTST Architecture}
\begin{figure}[htb]
   \centering
   \includegraphics[width=1.0\linewidth]{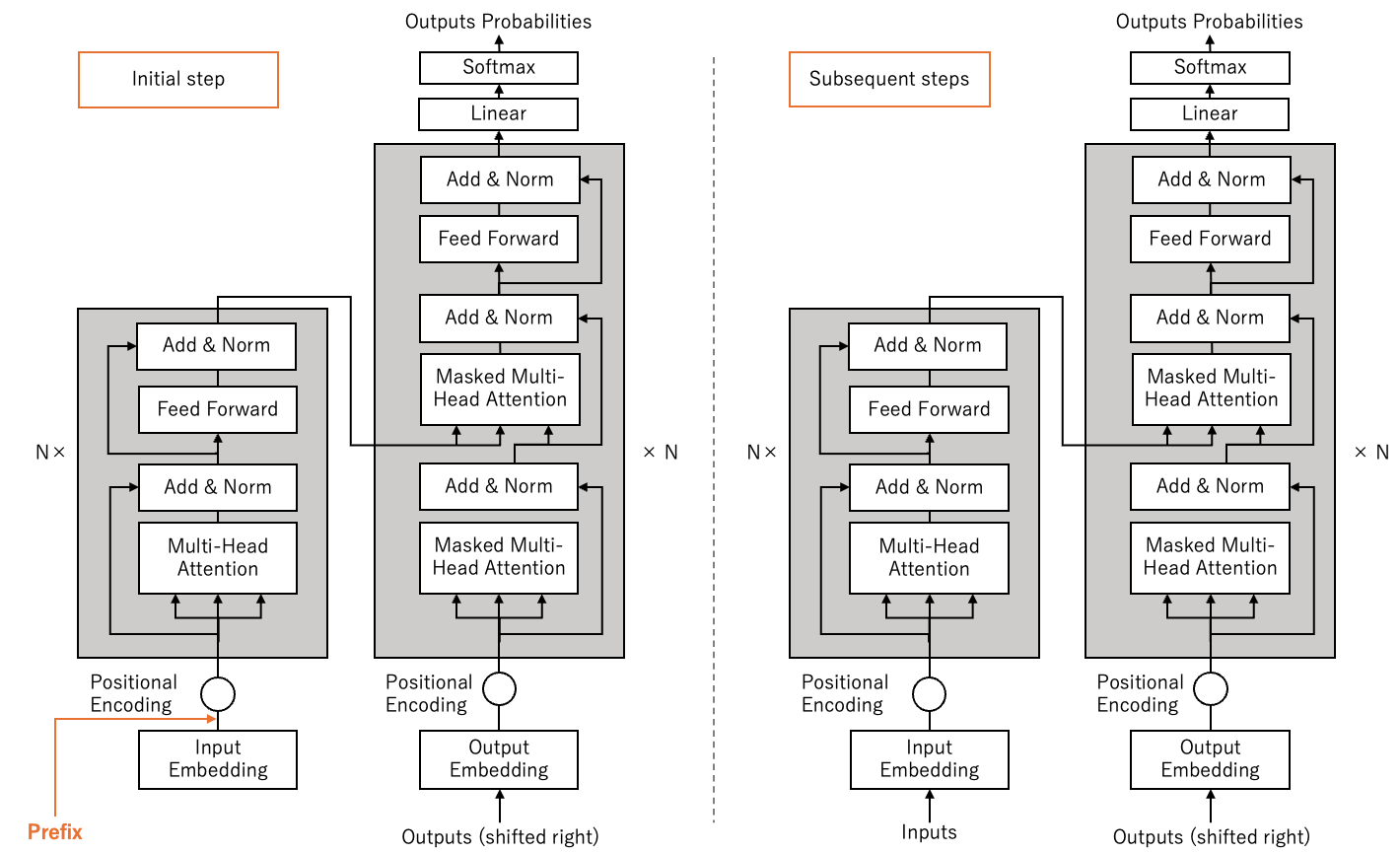}
   \caption{Architecture of Prefixed PatchTST model}
   \label{fig:architecture_prefix}
\end{figure}

\begin{figure}[htb]
\centering
\includegraphics[width=0.75\linewidth]{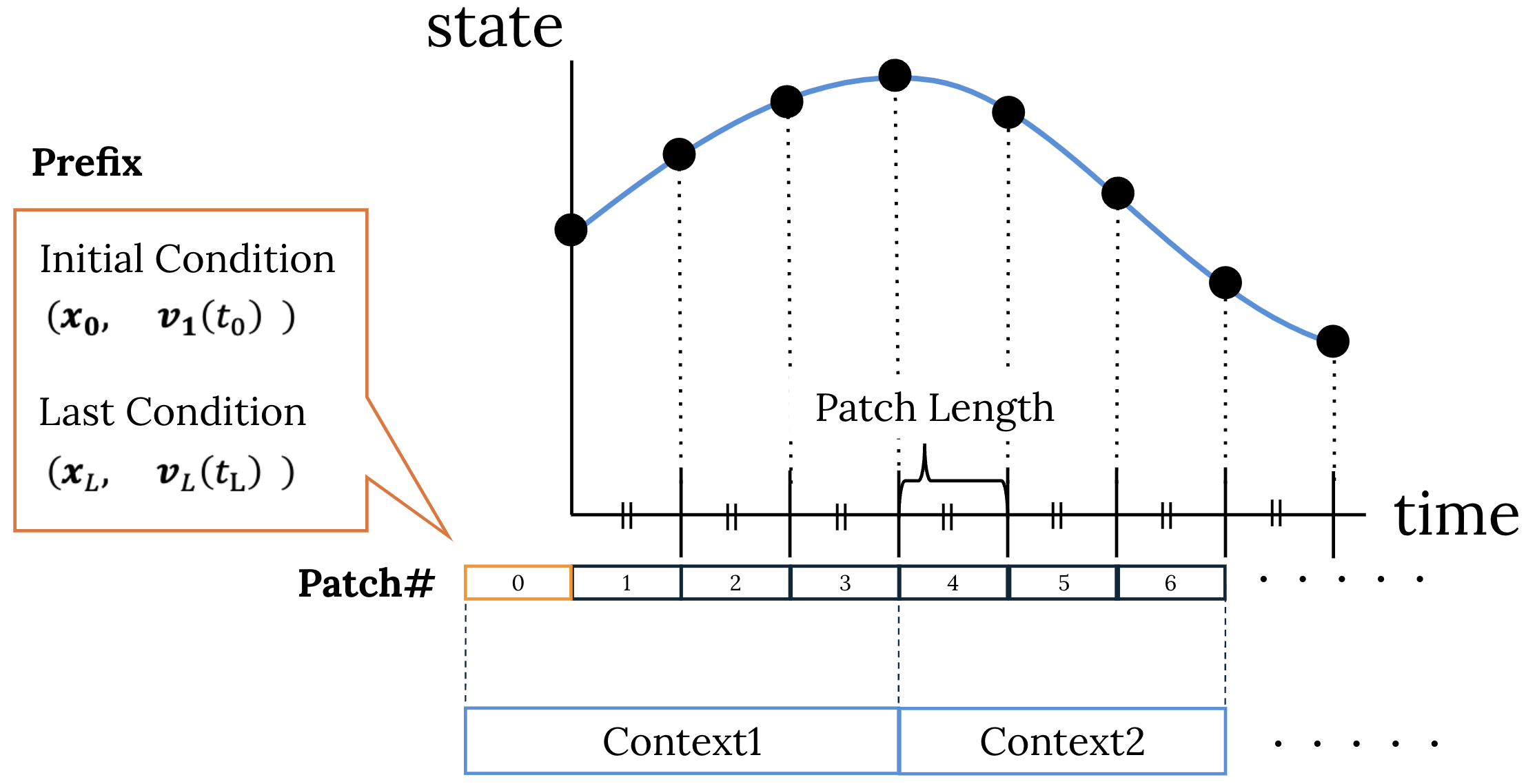}
\def\@captype{figure}
\figcaption{Overview of Prefixed PatchTST}
\label{fig:soft_prompt}
\end{figure}

We extend the standard PatchTST architecture by using prefix tokens, such as boundary conditions, to our model. As shown in Figure~\ref{fig:architecture_prefix}, the prefix tokens are directly connected to the Transformer Encoder in the first step, while in subsequent steps, trajectory data follows the standard processing path through Input Embedding and Positional Encoding. Prefix tokens provide critical information about initial and terminal position, allowing the model to generate trajectories that satisfy given boundary conditions, as shown in Figure~\ref{fig:soft_prompt}. In the first step, prefix tokens bypass the input embedding layer and proceed directly to positional encoding, while input patches go through both input embedding and positional encoding processes. For example, with a patch length of 8 and context length of 512, the input sequence consists of 64 patches that undergo input embedding, and one prefix token that skips this step. Both the prefix token and embedded patches then receive positional encoding. The prefix tokens serve as boundary conditions that control the transformer's output by providing constraints at initial and terminal positions. In subsequent steps, only the 64 patches with both input embeddings and positional encodings are processed through the transformer layers following the standard transformer architecture. Each patch retains both its embedding and positional information as it moves through the self-attention and feed-forward networks.

The separate processing path for prefix tokens enables the integration of boundary conditions while preserving the transformer's fundamental architecture. This approach minimizes architectural modifications to the existing transformer model, requiring changes only in the initial embedding stage, making it a highly reasonable solution for incorporating boundary value constraints.

\section{Numerical Simulation}
\subsection{Dataset Generation}
We generate trajectory datasets using trajectory propagation with different conditions for the lunar flyby in order to solve the two-point boundary value problem using Prefixed PatchTST, where the spacecraft is launched from Earth, passes near the Moon, and then travels to an arbitrary destination. The equations of motion, as described in equation \ref{EoM}, were numerically integrated using the DOP853 method with three different velocity configurations. The simulation parameters and generated dataset are detailed in Table \ref{tab:traj_params} and visualized in Figure~\ref{fig:dataset}, respectively.

The trajectory data has been preprocessed to be compatible with the input format of PatchTST. The forecast horizon is set to the multiple of patch length closest to 90 days - the time of flight for this problem. As a prefix to the proposed PatchTST, which is an input to this two-point boundary value problem, we extract the initial and terminal positions from the dataset.
As an initial context length for PatchTST, we provide a trajectory that is back-propagated from the lunar flyby to Earth. For this lunar flyby, we assume a zero sphere of influence (SOI) Patched Conics method, where the incoming v-infinity is uniquely determined regardless of the outgoing v-infinity value for back-propagation. The incoming v-infinity is selected through a grid search, varying both magnitude and direction, to find trajectories that approach Earth. The outgoing v-infinity conditions are adjusted to ensure that the lunar flyby altitude constraints are satisfied.

\begin{table}[htb]
\begin{minipage}[t]{0.5\textwidth}
  \centering
  \begin{table}[H]
  \caption{Trajectory Generation Parameters}
  \begin{tabular}{lll}
      \hline
      Parameter & Value & Units \\
      \hline
      \multicolumn{3}{l}{\textit{Initial Conditions}} \\
      Initial position & Moon & - \\
      Approach angle & 0 & degree \\
      $V_{\infty}$ conditions & 1.2, 1.3, 1.4 & km/s \\
      Post-flyby angle & [-90, 90] & degree \\
      \hline
      \multicolumn{3}{l}{\textit{Trajectory Generation}} \\
      Trajectories per $V_{\infty}$ & 1,000 & - \\
      Total trajectories & 3,000 & - \\
      Forward propagation & 90 & days \\
      Backward propagation & 512 & steps \\
      Sampling interval & 7 & minute \\
      \hline
      \multicolumn{3}{l}{\textit{Data Split}} \\
      Training data & 70 & \% \\
      Validation data & 10 & \% \\
      Test data & 20 & \% \\
      \hline
  \end{tabular}
  \label{tab:traj_params}
  \end{table}
\end{minipage}
\hfill
\begin{minipage}[t]{0.5\textwidth}
 \centering
 \begin{figure}[H]
 \includegraphics[width=1.0\linewidth]{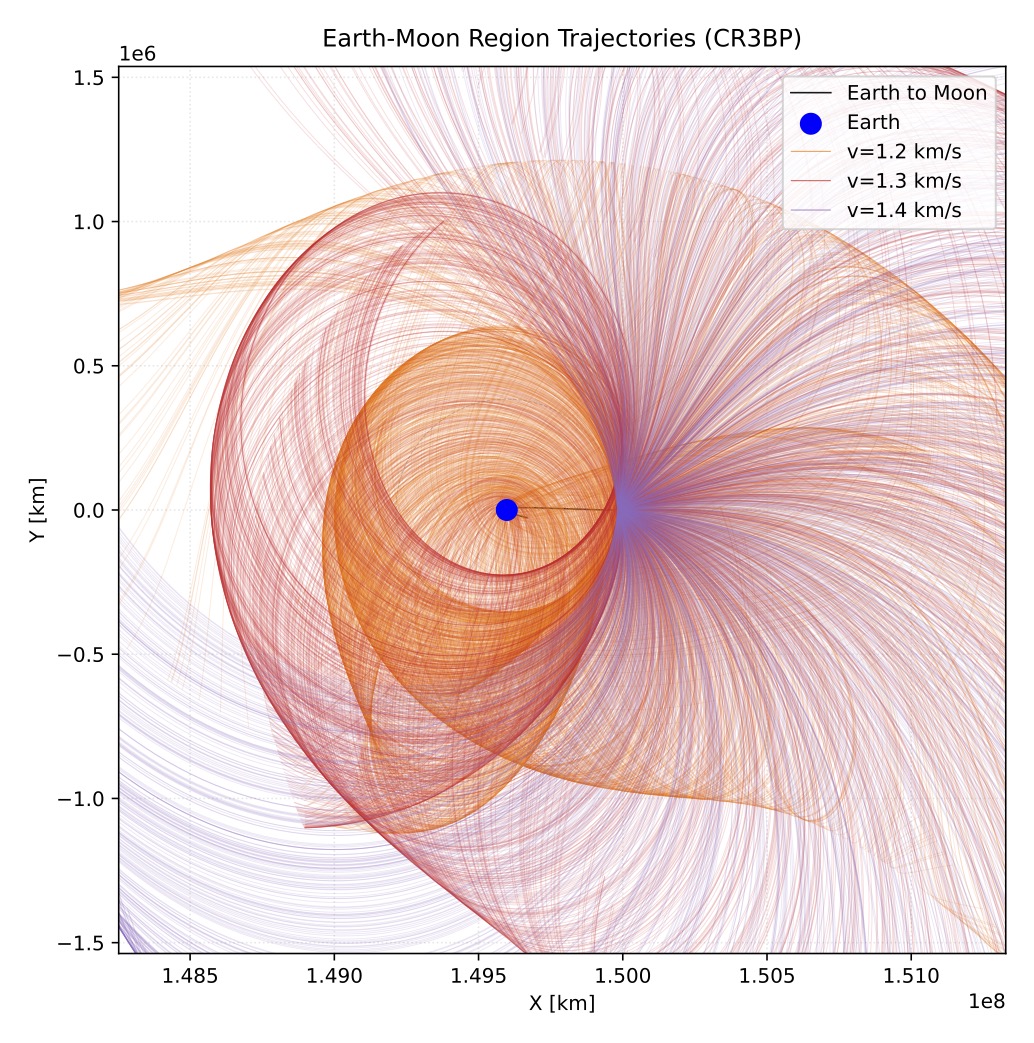}
 \caption{Generated dataset}
 \label{fig:dataset}
 \end{figure}
\end{minipage}
\end{table}

\subsection{Training Process and Hyperparameter Tuning}
We first tune the hyperparameters with reduced epochs, and then train with optimized parameters using the full epoch setting. 

\begin{table}[h]
\centering
\caption{Model Configuration Parameters}
\begin{tabular}{ll}
\hline
Parameter & Value \\
\hline
Input Channels & Variable (based on forecast columns) \\
Context Length & 512 \\
Patch Length & 16 \\
d\_model & 128 \\
Number of Attention Heads & 16 \\
Number of Hidden Layers & Variable (6 by default) \\
FFN Dimension & Variable (256 by default) \\
Dropout Rate & 0.2 \\
Head Dropout & 0.2 \\
Loss Function & MSE \\
Optimizer & Adam optimizer \\
Evaluation Strategy & every 1\% of training steps \\
ffn dim         & 768  \\
shuffle seed    & 123  \\
 \hline
\end{tabular}
\end{table}

\begin{figure}[htb]
    \centering
    \includegraphics[width=\linewidth]{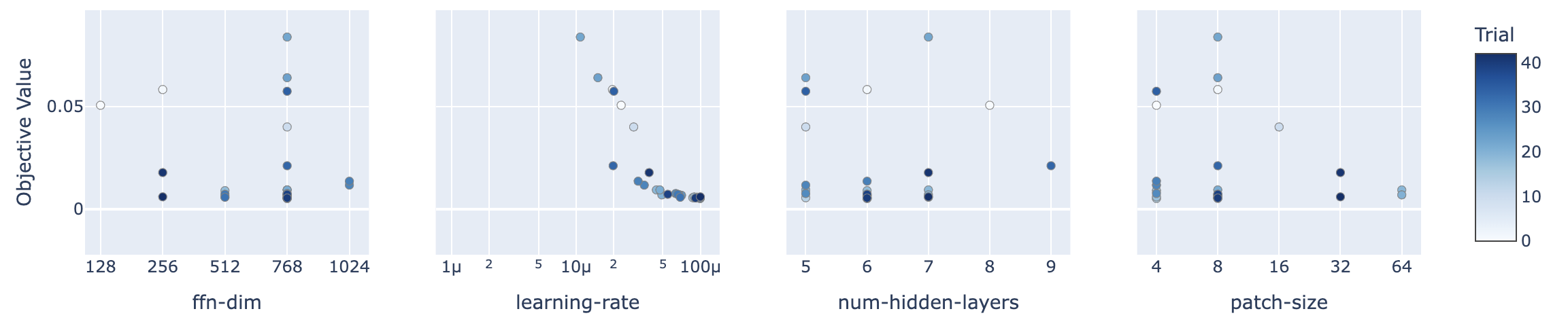}
    \caption{Hyperparameter Optimization Results}
    \label{fig:result_optuna}
\end{figure}

We conducted comprehensive hyperparameter tuning using Optuna \cite{optuna_2019}, which implements the Tree-structured Parzen Estimator (TPE) \cite{NIPS2011_86e8f7ab}, a Bayesian optimization approach. During the tuning process, the model was trained with different parameter combinations, and the loss was calculated based on the position error of the generated trajectories. The optimization process explored the search space defined in Table \ref{tab:optuna-ranges}, which includes critical parameters such as patch length, number of hidden layers, FFN dimension, and learning rate. Each trial involved training the model using PyTorch and evaluating its performance based on the state prediction error. Figure~\ref{fig:result_optuna} illustrates the results of the hyperparameter optimization, showing the relationship between different parameter values on the horizontal axis and their corresponding loss values on the vertical axis. The intensity of the color indicates the progression of the trials, and darker colors represent later trials.

Based on the optimization results, we identified the optimal hyperparameter configuration shown in Table \ref{tab:optuna}. The learning rate demonstrated particularly significant impact on model performance, leading us to select $5e-4$ as the optimal value. We made practical adjustments to certain parameters: the Context Length was set to 512 to accommodate Earth-Moon orbit durations, while the Future Horizon was configured to 17,984 to enable 90-day sequence generation. Following the hyperparameter optimization phase, we proceeded with the full training process using the optimized architecture. The model was trained for an extended period of 16,000 epochs to ensure convergence and optimal performance in trajectory generation. This comprehensive approach to hyperparameter tuning and training resulted in a model capable of generating accurate and stable orbital trajectories.

\begin{table}[h]
    \centering
    \fontsize{10}{10}\selectfont
    \caption{Search Space of Hyperparameter Optimization}
    \begin{tabular}{lll}
        \hline
        Hyperparameter & Range or Choices & Type \\
        \hline
        patch-size & \{4, 8, 16, 32, 64\} & Categorical \\
        hidden-layers & [5, 9] & Integer \\
        ffn-dim & \{128, 256, 512, 768, 1024\} & Categorical \\
        learning-rate & [$10^{-6}$, $10^{-4}$] & Float (log scale) \\
        \hline
    \end{tabular}
    \label{tab:optuna-ranges}
\end{table}

\begin{table}[h]
    \centering
    \caption{Optimized Hyperparameters}
    \label{tab:optuna}
    \begin{tabular}{c | r }
        \hline 
        Parameter    & Value \\
        \hline 
        Layers            & 8 \\
        Epochs            & 16000 \\
        Context Length    & 512 \\
        Patch length    & 32 \\
        Learning rate   & 5e-4 \\
        Forecast Horizon   & 17984 \\
        \hline
    \end{tabular}
\end{table}

\subsection{Performance of Prefixed PatchTST}
\begin{figure}[htb]
  \centering
  \includegraphics[width=1.0\linewidth]{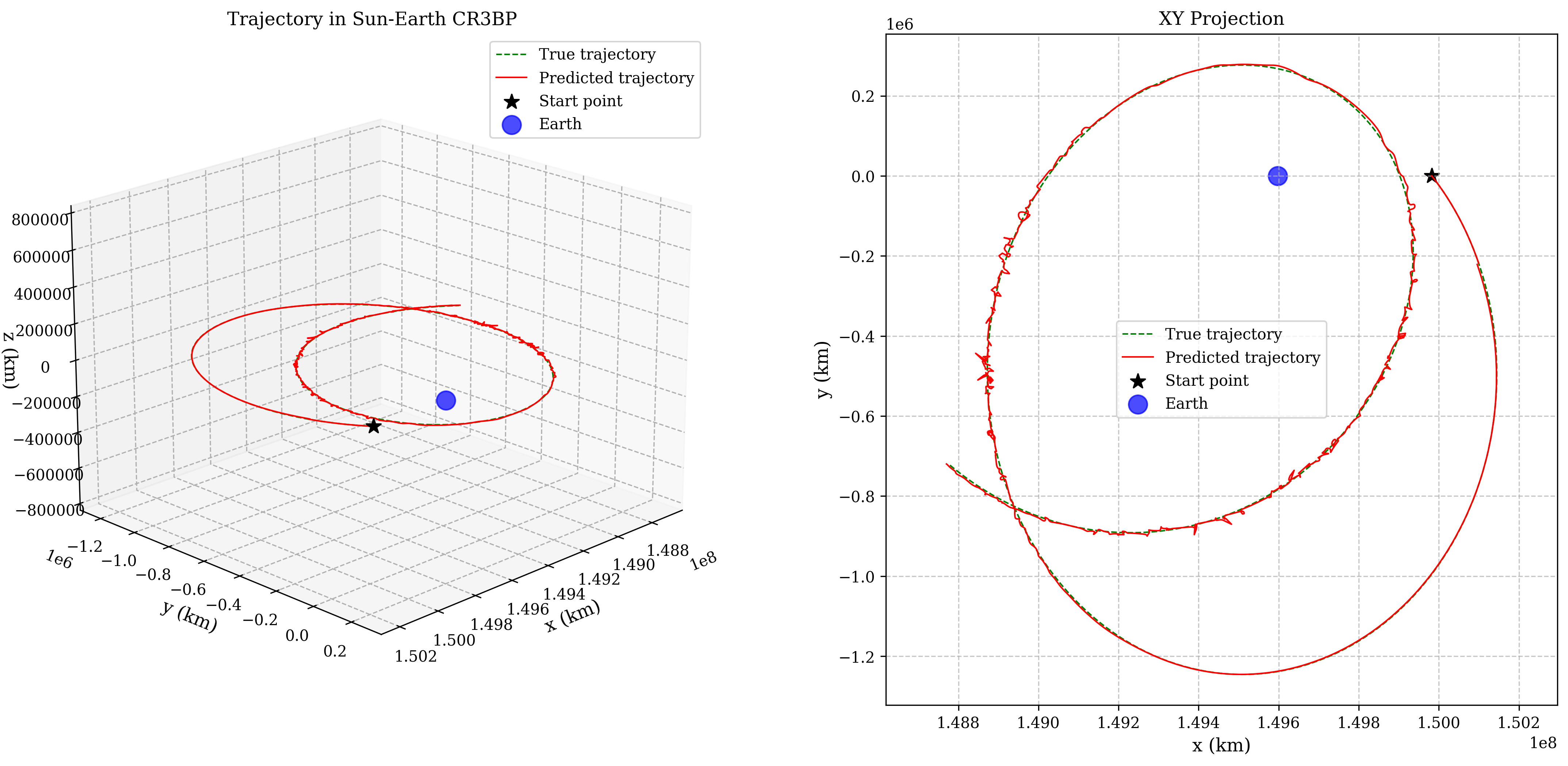}
  \caption{Three-dimensional comparison of generated and ground truth trajectories}
  \label{fig:3d_trajectory}
\end{figure}

\begin{figure}[htb]
  \centering
  \includegraphics[width=1.0\linewidth]{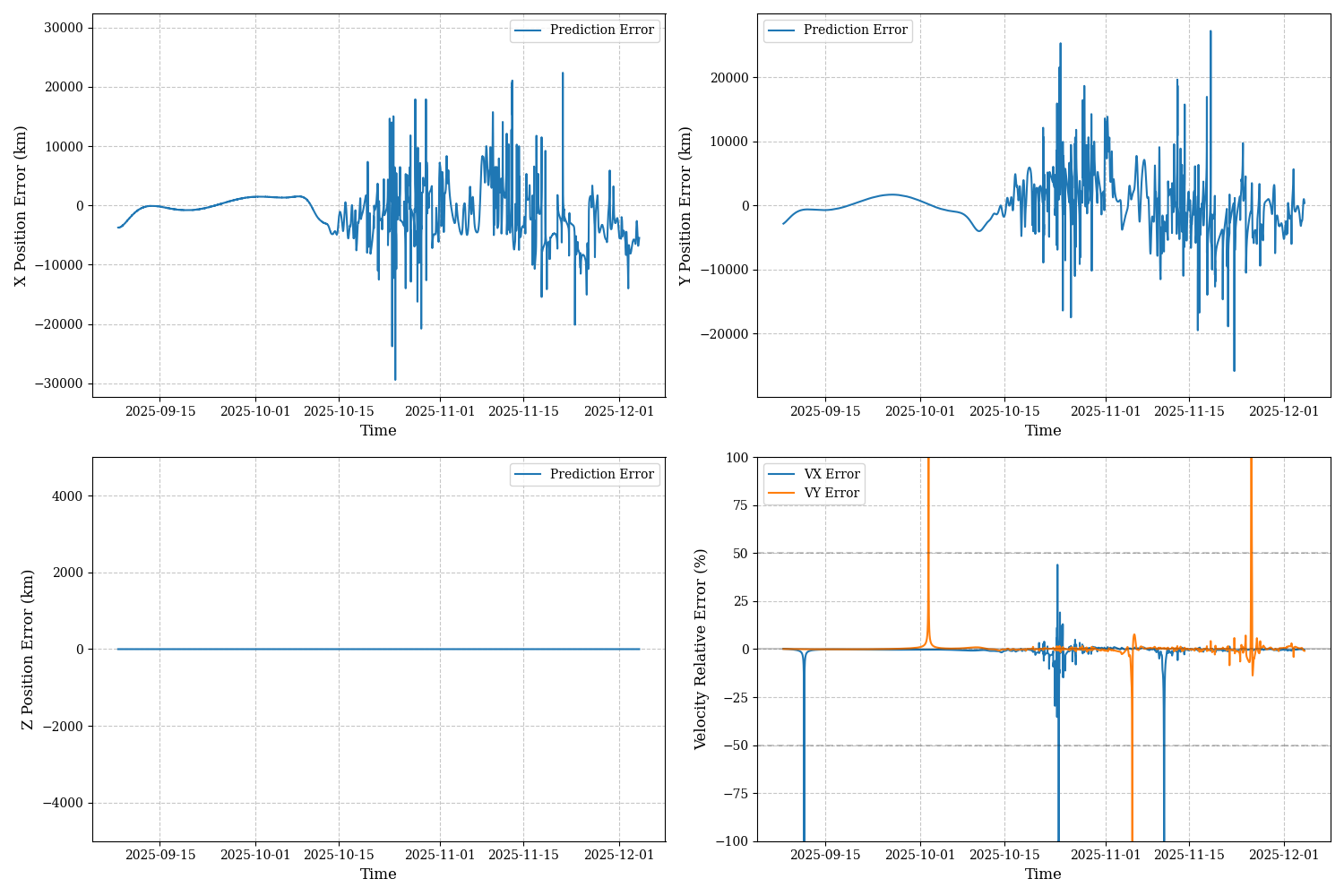}
  \caption{Error time evolution of state variables}
  \label{fig:state_evolution}
\end{figure}

We evaluated the performance of our model by comparing the generated trajectories with the true trajectory obtained from the numerical integration of the equations of motion. Figure~\ref{fig:3d_trajectory} shows trajectories generated by our Prefixed PatchTST model and the true trajectory. The initial and terminal points are conditional on the input, and the red trajectory is the output, which is the solution to the two-point boundary value problem that PatchTST entirely inferred by using that input information. The inferred trajectory has reached the terminal point, and the trajectory between the two trajectories is generally moving along the green true trajectory. The trajectory inferred by the generative model is dynamically unrealistic, but it captures the overall trend. Figure~\ref{fig:state_evolution} shows the evolution of the states' errors with respect to time. More detailed results are provided in the Appendix. Position errors remaining below approximately 10,000 km and $z$ component of both position and velocity show strong agreement with the true data, frequently achieving near-zero errors as shown in Figure~\ref{fig:state_evolution}, since the training data we are giving in this training only has a $z=0$.

Although the model demonstrates high accuracy in trajectory prediction as shown above, the generated trajectories exhibit intermittent fluctuations. These intermittent patterns likely stem from the inherent limitations in the transformer model's representation capability and the finite size of our training dataset. There is a zigzag in the inferred trajectory, but on a global scale, it is consistent with the true trajectory, so it should be usable as an initial predicted trajectory for trajectory optimization. The occurrence of these oscillations suggests that the model may benefit from additional training data or architectural modifications to better capture long-term dependencies in the trajectory dynamics. Despite these oscillations, our prefixed PatchTST model effectively learns the underlying dynamics of lunar flyby trajectories. A comprehensive set of generated trajectories is presented in the Appendix.

\subsection{Statistical Analysis of Prefixed PatchTST Prediction}
\begin{figure}[htb]
\centering
\includegraphics[width=1.0\linewidth]{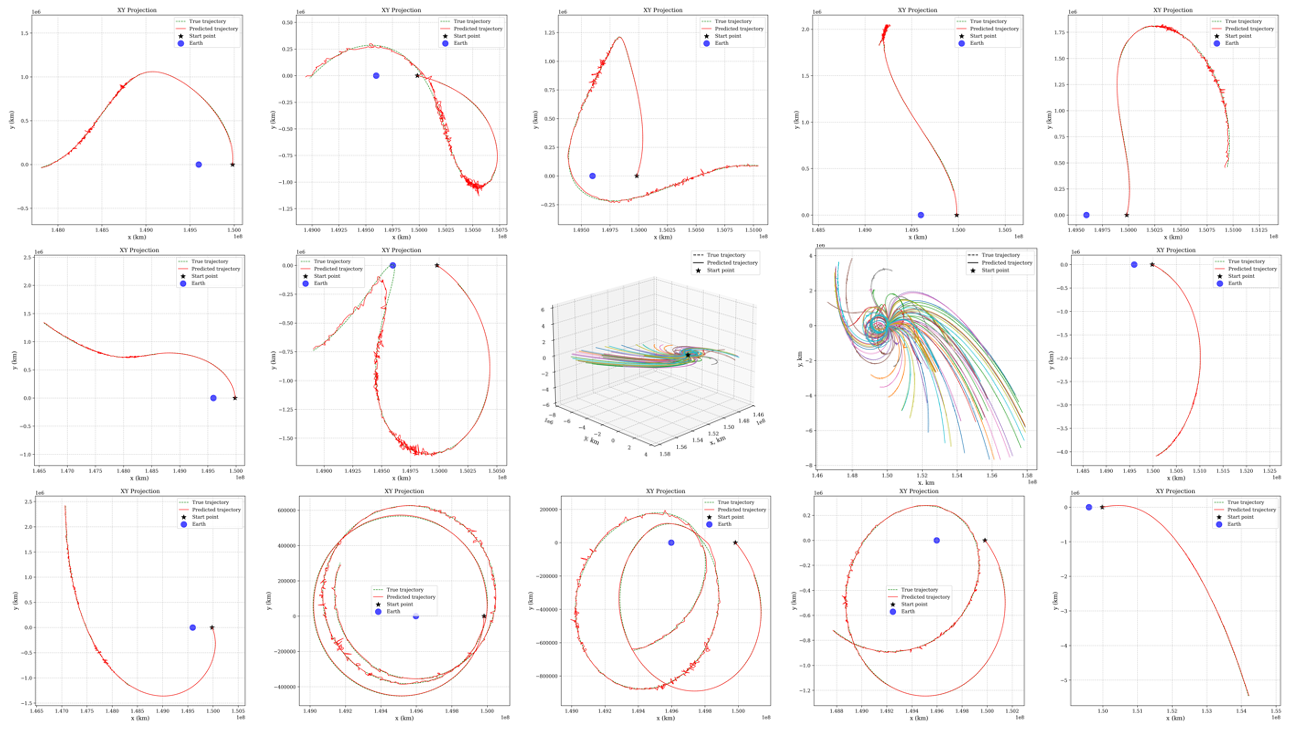}
\caption{Three-dimensional comparison of 100 generated trajectories (solid lines) and true trajectories (dashed lines)}
\label{fig:100_traj}
\end{figure}

To analyze the performance more quantitatively, we conducted a statistical analysis of Prefixed PatchTST prediction. The analysis assesses 100 trajectories with different prefix conditions to visualize the behavior of the model. Figure~\ref{fig:100_traj} shows the comparison between the generated trajectories and the true data of 13 representative cases, which includes escape orbit via L1, L2, Earth flyby orbits, and multi-revolution orbit. Even with these trajectories, Prefixed PatchTST generally exhibited good accuracy. In the Earth flyby, we can observe the significant discrepancies between predicted and actual orbits. This is due to the high sensitivity of the incoming trajectory, making its hyperbolic orbit difficult to predict. The sensitivity to the incoming trajectory is high, making it difficult to predict its hyperbolic orbit. However, despite initial inaccuracies, the final predicted location aligns closely with the actual trajectory, demonstrating improved precision as the object approaches its destination.

\begin{figure}[htb]
\centering
\includegraphics[width=1.0\linewidth]{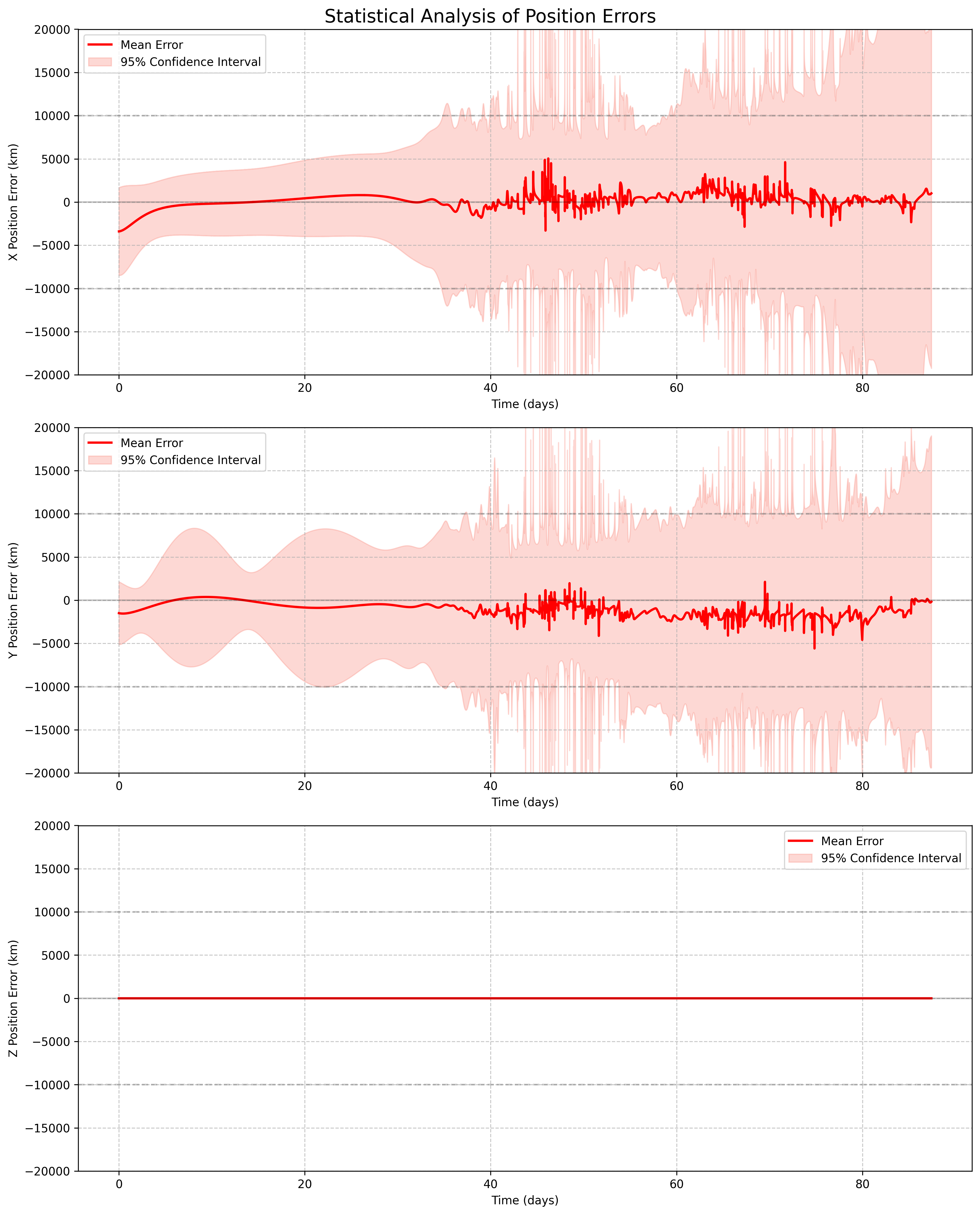}
\caption{Statistical analysis of position errors over 90 days showing mean error (red line) and 95\% confidence interval (shaded area) for x, y, and z components}
\label{fig:position_error}
\end{figure}

\begin{figure}[htb]
\centering
\includegraphics[width=1.0\linewidth]{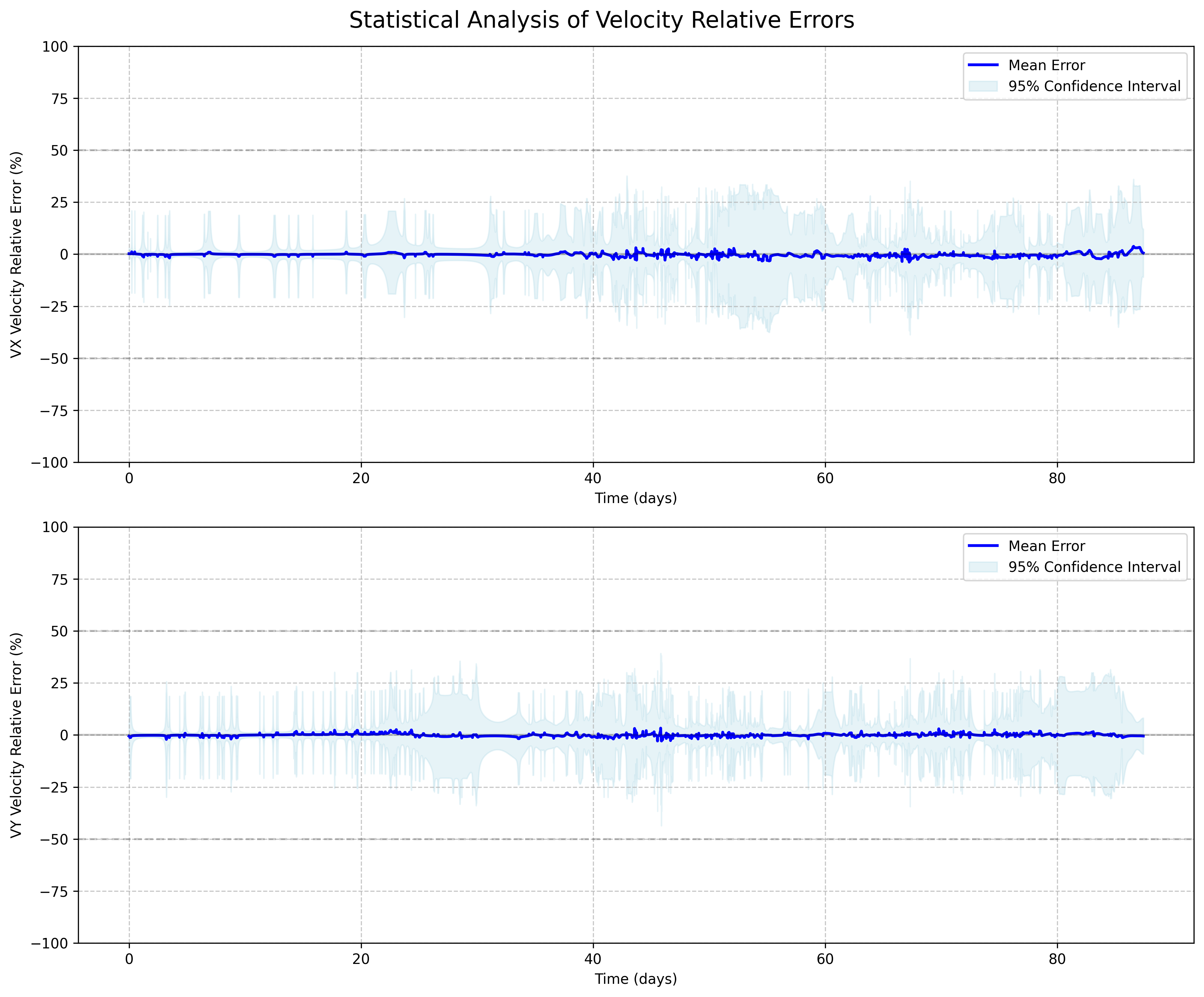}
\caption{Statistical analysis of velocity errors over 90 days showing mean error (blue line) and 95\% confidence interval (shaded area) for Vx, Vy, and Vz components}
\label{fig:velocity_error}
\end{figure}

Figures~\ref{fig:position_error} and \ref{fig:velocity_error} show the statistical distribution of positions and velocities over time. 
The red line shows the mean error and the pink region describes the 95\% confidence interval for each component over 90 days.
The analysis of both position and velocity errors shows that the 95\% confidence intervals widen over time. The $z$ component of both position and velocity maintains relatively stable error bounds, frequently achieving near-zero errors.

\section{Conclusions}
This study has demonstrated the potential of prefixed PatchTST for solving two-point boundary value problems in three-body dynamics. By incorporating boundary conditions as prefix tokens, our model has successfully generated trajectories that connect specified initial and terminal positions in the CR3BP. PatchTST with the introduction of Prefix showed good accuracy, which is also an important contribution in the field of Machine Learning. Statistical analysis of the proposed method revealed that our prefix approach performs well for a wide variety of trajectories, including escape orbit via L1, L2, Earth flyby orbits, and multi-revolution orbit.

\section{Acknowledgment}
This work was supported by G-7 Foundation Research and Development Grant Program (Biotechnology and IT fields).

\bibliographystyle{AAS_publication}   
\bibliography{references}   

\begin{thebibliography}{10}

\bibitem{ken2020}
K.~Oguri, K.~Oshima, S.~Campagnola, K.~Kakihara, N.~Ozaki, N.~Baresi, Y.~Kawakatsu, and R.~Funase, ``EQUULEUS Trajectory Design.,''  {\em The Journal of the Astronautical Sciences}, Vol.~67, 2020, p.~950–976.

\bibitem{Fe2024}
F.~L. and Z.~A., ``Robust interplanetary trajectory design under multiple uncertainties via meta-reinforcement learning,''  {\em Acta Astronautica}, Vol.~214, 2023, pp.~147–--158.

\bibitem{sullivan2020using}
C.~J. Sullivan and N.~Bosanac, ``Using reinforcement learning to design a low-thrust approach into a periodic orbit in a multi-body system,''  {\em AIAA scitech 2020 forum}, 2020, p.~1914.

\bibitem{Zavoli2021reinforcement}
A.~Zavoli and L.~Federici, ``Reinforcement learning for robust trajectory design of interplanetary missions,''  {\em Journal of Guidance, Control, and Dynamics}, Vol.~44, No.~8, 2021, pp.~1440--1453.

\bibitem{li2023amortized}
A.~Li, A.~Sinha, and R.~Beeson, ``Amortized Global Search for Efficient Preliminary Trajectory Design with Deep Generative Models,''  {\em arXiv preprint arXiv:2308.03960}, 2023.

\bibitem{Izzo2021}
D.~Izzo and E.~Öztürk, ``Real-time guidance for low-thrust transfers using deep neural networks,''  {\em Journal of Guidance, Control, and Dynamics}, Vol.~44, No.~2, 2021, pp.~315–--327.

\bibitem{Guffanti2024transformers}
T.~Guffanti, D.~Gammelli, S.~D’Amico, and M.~Pavone, ``Transformers for Trajectory Optimization with Application to Spacecraft Rendezvous,''  {\em 2024 IEEE Aerospace Conference}, IEEE, 2024, pp.~1--13.

\bibitem{Presser2024}
T.~Presser, A.~Dasgupta, D.~Erwin, and A.~Oberai, ``Diffusion Models for Generating Ballistic Spacecraft Trajectories,''  {\em 2024 Astrodynamics Specialist Conference}, 2024.

\bibitem{briden2025diffusion}
J.~Briden, B.~J. Johnson, R.~Linares, and A.~Cauligi, ``Diffusion Policies for Generative Modeling of Spacecraft Trajectories,''  {\em AIAA SCITECH 2025 Forum}, 2025, p.~2775.

\bibitem{box2015time}
G.~E. Box, G.~M. Jenkins, G.~C. Reinsel, and G.~M. Ljung, {\em Time series analysis: forecasting and control}.
\newblock John Wiley \& Sons, 2015.

\bibitem{McKenzie1984}
E.~McKenzie, ``General exponential smoothing and the equivalent ARMA process,''  {\em Journal of Forecasting}, Vol.~3, No.~3, 1984, pp.~333--344.

\bibitem{NIPS2017_3f5ee243}
A.~Vaswani, N.~Shazeer, N.~Parmar, J.~Uszkoreit, L.~Jones, A.~N. Gomez, L.~u. Kaiser, and I.~Polosukhin, ``Attention is All you Need,''  {\em Advances in Neural Information Processing Systems} (I.~Guyon, U.~V. Luxburg, S.~Bengio, H.~Wallach, R.~Fergus, S.~Vishwanathan, and R.~Garnett, eds.), Vol.~30, Curran Associates, Inc., 2017.

\bibitem{oreshkin2019n}
B.~N. Oreshkin, D.~Carpov, N.~Chapados, and Y.~Bengio, ``N-BEATS: Neural basis expansion analysis for interpretable time series forecasting,''  {\em arXiv preprint arXiv:1905.10437}, 2019.

\bibitem{das2023decoder}
A.~Das, W.~Kong, R.~Sen, and Y.~Zhou, ``A decoder-only foundation model for time-series forecasting,''  {\em arXiv preprint arXiv:2310.10688}, 2023.

\bibitem{Nie2022}
Y.~Nie, N.~H. Nguyen, P.~Sinthong, and J.~Kalagnanam, ``A Time Series is Worth 64 Words: Long-term Forecasting with Transformers,''  {\em Arxiv}, 2022.

\bibitem{kenton2019bert}
J.~D. M.-W.~C. Kenton and L.~K. Toutanova, ``Bert: Pre-training of deep bidirectional transformers for language understanding,''  {\em Proceedings of naacL-HLT}, Vol.~1, Minneapolis, Minnesota, 2019, p.~2.

\bibitem{brown2020language}
T.~Brown, B.~Mann, N.~Ryder, M.~Subbiah, J.~D. Kaplan, P.~Dhariwal, A.~Neelakantan, P.~Shyam, G.~Sastry, A.~Askell, {\em et~al.}, ``Language models are few-shot learners,''  {\em Advances in neural information processing systems}, Vol.~33, 2020, pp.~1877--1901.

\bibitem{optuna_2019}
T.~Akiba, S.~Sano, T.~Yanase, T.~Ohta, and M.~Koyama, ``Optuna: A Next-generation Hyperparameter Optimization Framework,''  {\em Proceedings of the 25th {ACM} {SIGKDD} International Conference on Knowledge Discovery and Data Mining}, 2019.

\bibitem{NIPS2011_86e8f7ab}
J.~Bergstra, R.~Bardenet, Y.~Bengio, and B.~K\'{e}gl, ``Algorithms for Hyper-Parameter Optimization,''  {\em Advances in Neural Information Processing Systems} (J.~Shawe-Taylor, R.~Zemel, P.~Bartlett, F.~Pereira, and K.~Weinberger, eds.), Vol.~24, Curran Associates, Inc., 2011.

\end{thebibliography}

\clearpage
\newpage
\section{Appendix}
\subsection{Lunar Flyby Constraints}
The angle between incoming $v_{inf}$ and outgoing $v_{inf}$ is defined as deflection angle. The deflection angle $\phi_B$ of the flyby trajectory can be theoretically calculated using:

\begin{align}
\sin{\dfrac{\phi_B}{2}} = \dfrac{1}{1+\dfrac{r_{\pi}V_{\infty}^2}{\mu}} \qquad (0^{\circ}\leq \phi_B \leq 180^{\circ})
\end{align}
where $r_{\pi}$ is the periapse radius, $V_{\infty}$ is the hyperbolic excess velocity, and $\mu$ is the Moon's gravitational parameter. By constraining the minimum periapse radius $r_{\pi}$, we can effectively limit the maximum turning angle $\phi_B$.

\subsection{Comprehensive Trajectories Generated by Prefixed PatchTST}
\begin{figure}[htb]
  \centering
  \includegraphics[width=1.0\linewidth]{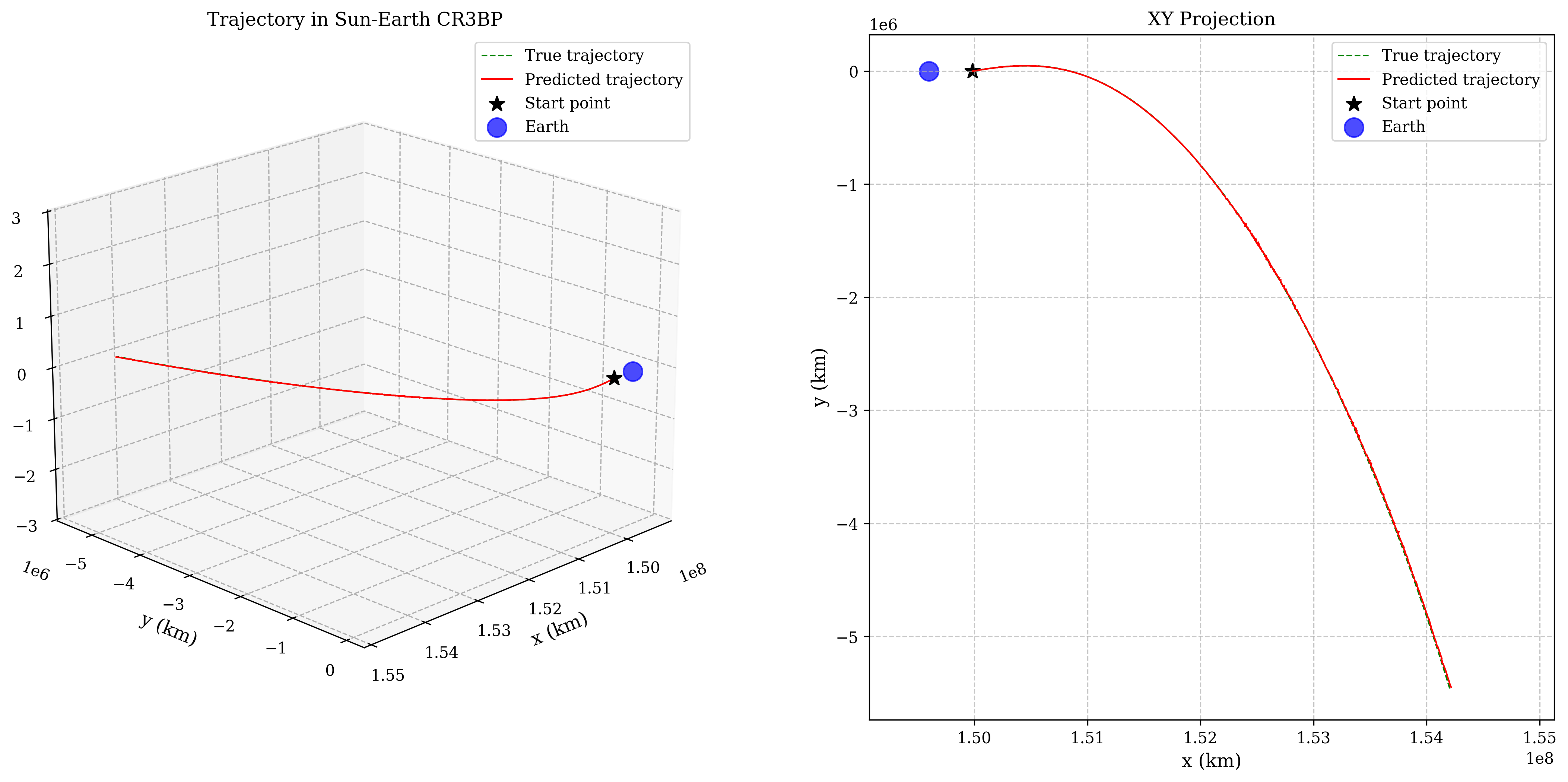}
  \caption{Three-dimensional comparison of generated and ground truth trajectories (Case 1)}
  \label{fig:3d_trajectory1}
\end{figure}

\begin{figure}[htb]
  \centering
  \includegraphics[width=1.0\linewidth]{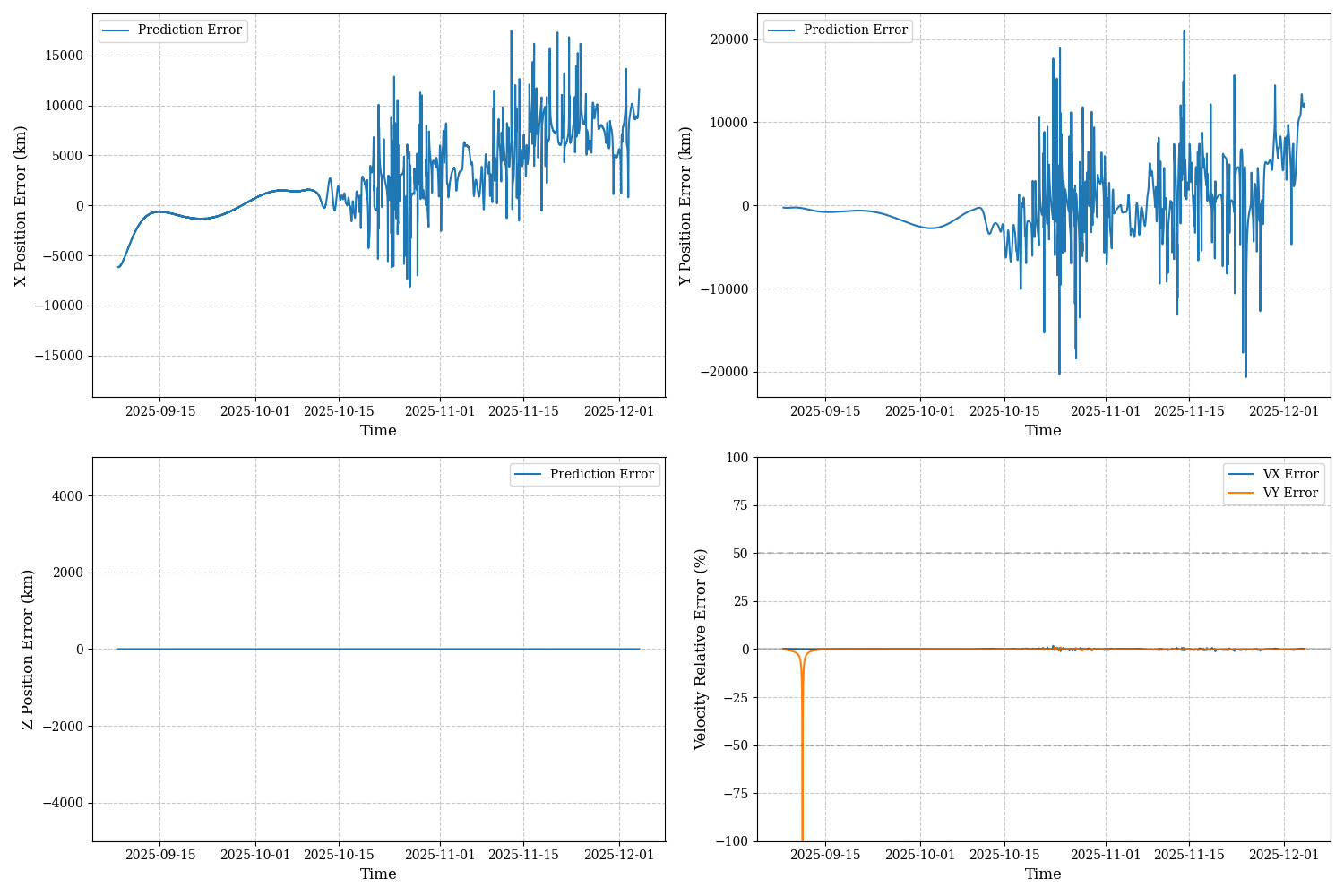}
  \caption{Error time evolution of state variables (Case 1)}
  \label{fig:state_evolution1}
\end{figure}

\begin{figure}[htb]
  \centering
  \includegraphics[width=1.0\linewidth]{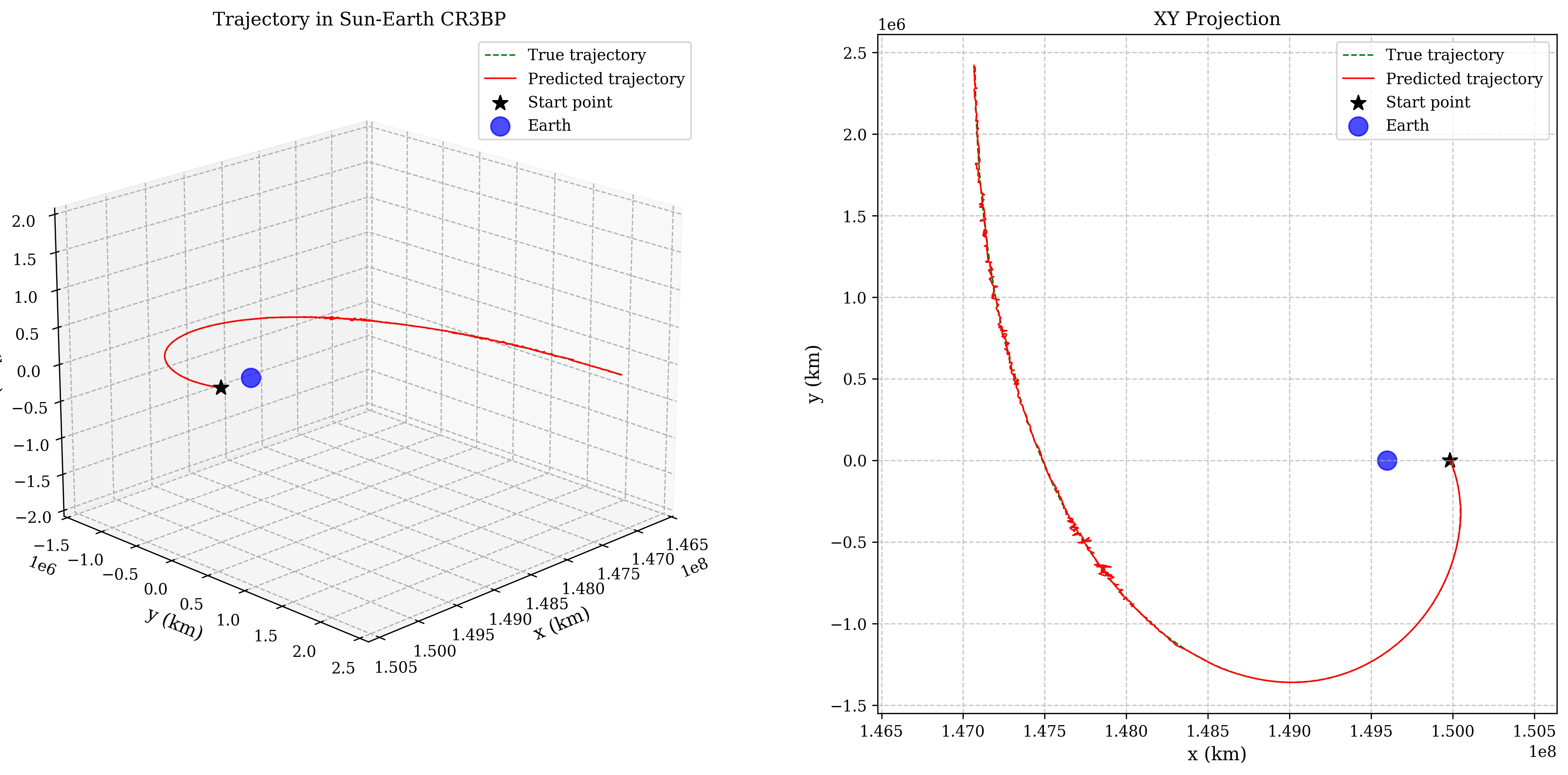}
  \caption{Three-dimensional comparison of generated and ground truth trajectories (Case 2)}
  \label{fig:3d_trajectory9}
\end{figure}

\begin{figure}[htb]
  \centering
  \includegraphics[width=1.0\linewidth]{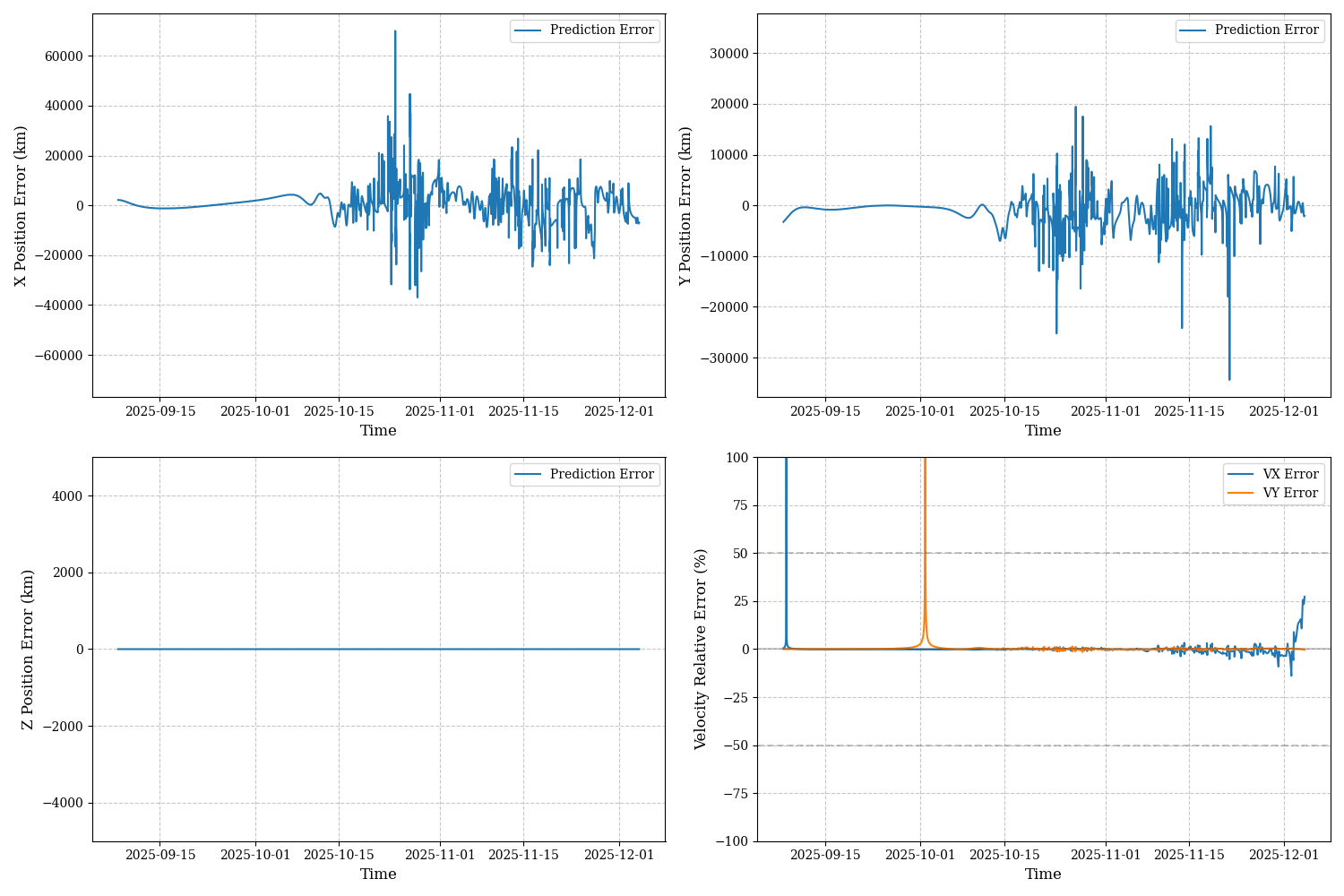}
  \caption{Error time evolution of state variables (Case 2)}
  \label{fig:state_evolution9}
\end{figure}

\begin{figure}[htb]
  \centering
  \includegraphics[width=1.0\linewidth]{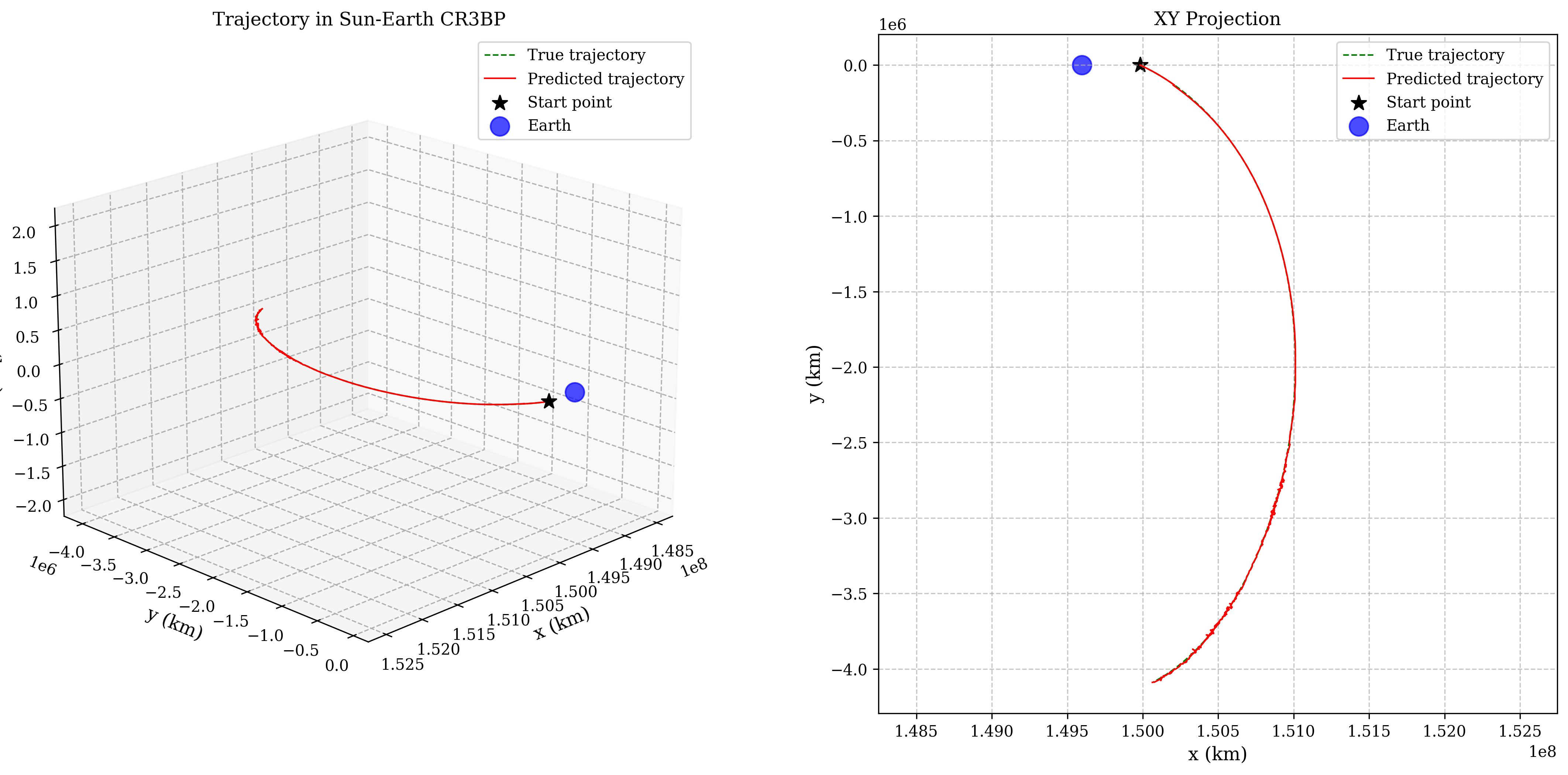}
  \caption{Three-dimensional comparison of generated and ground truth trajectories (Case 3)}
  \label{fig:3d_trajectory11}
\end{figure}

\begin{figure}[htb]
  \centering
  \includegraphics[width=1.0\linewidth]{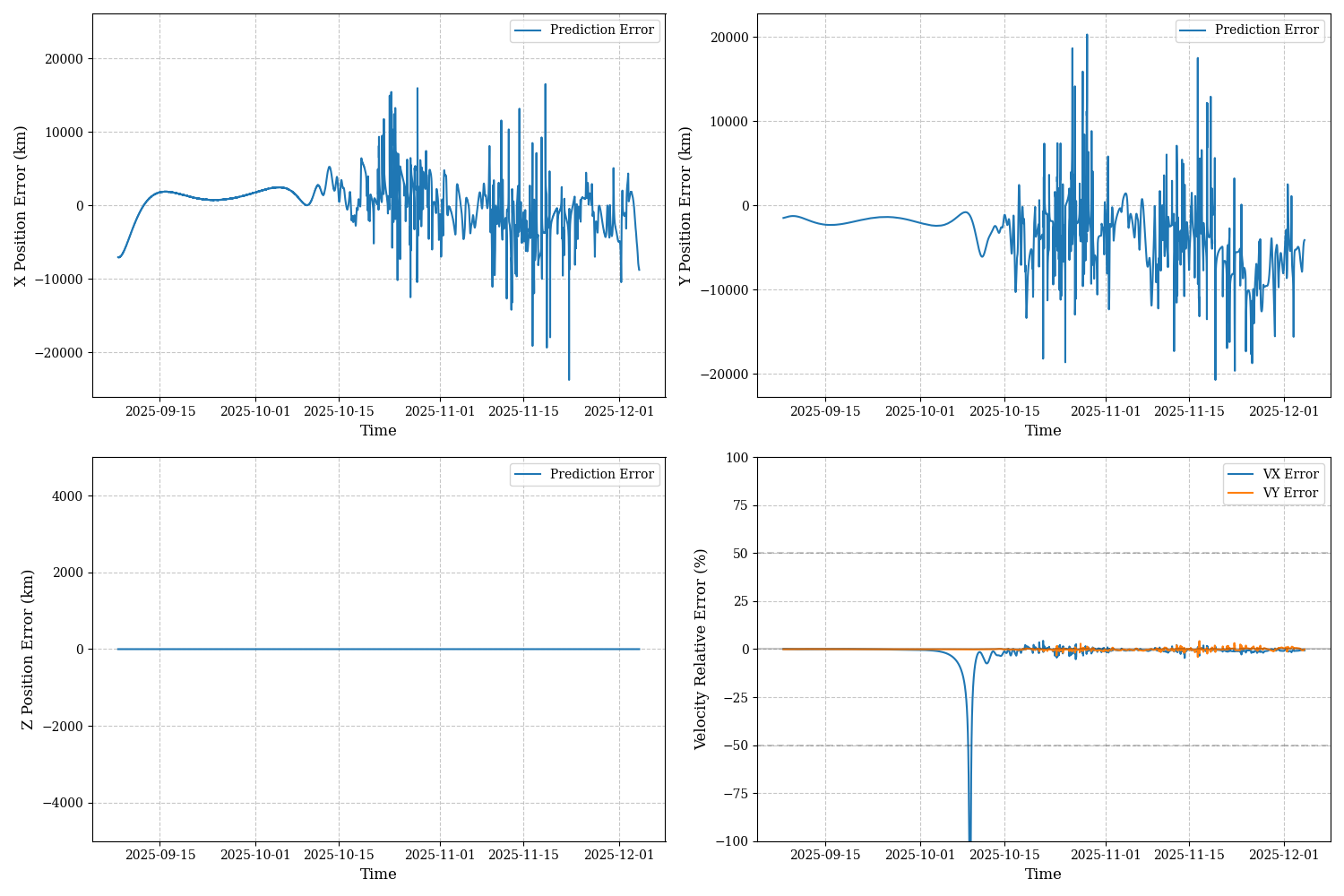}
  \caption{Error time evolution of state variables (Case 3)}
  \label{fig:state_evolution11}
\end{figure}

\begin{figure}[htb]
  \centering
  \includegraphics[width=1.0\linewidth]{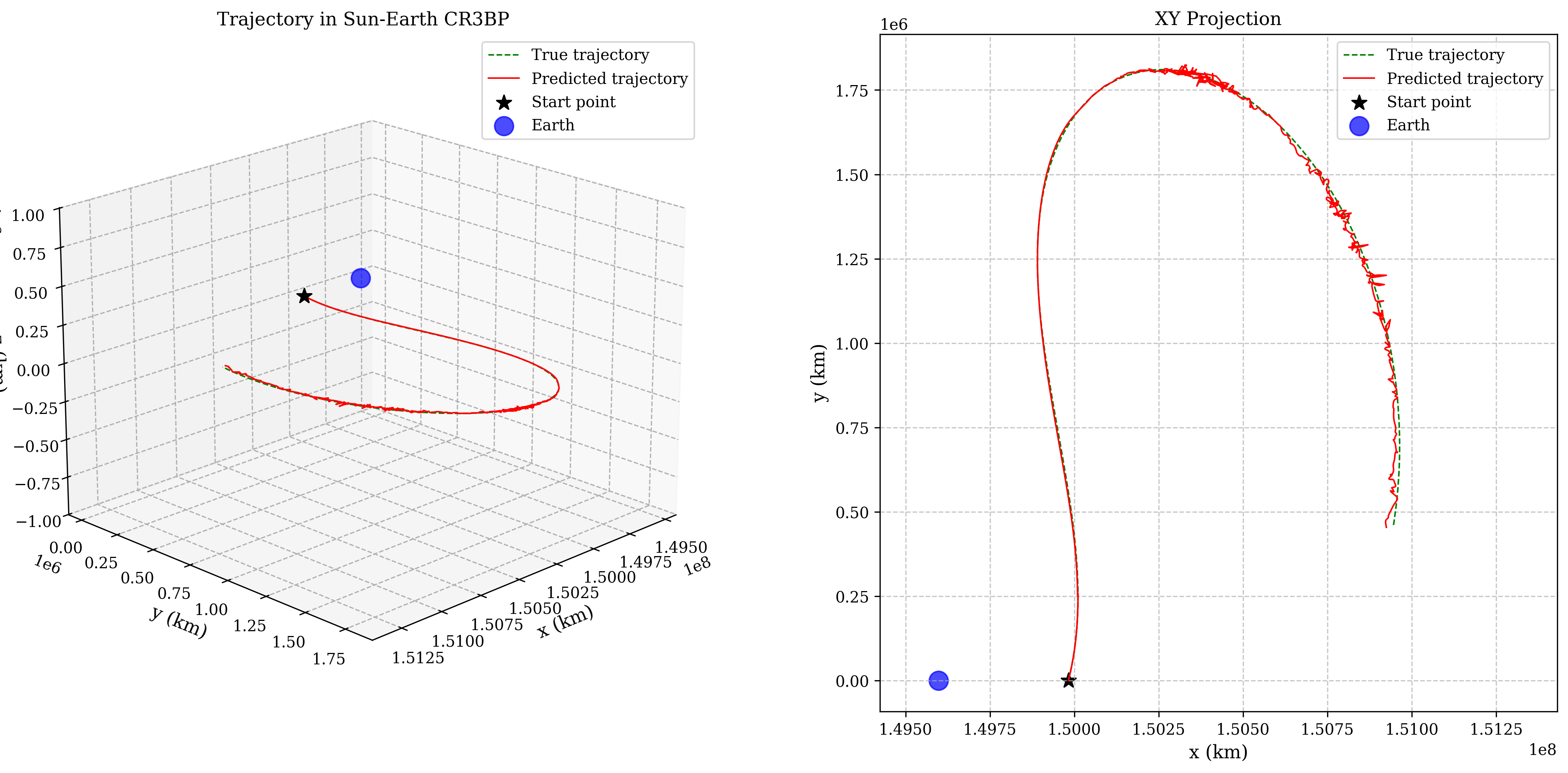}
  \caption{Three-dimensional comparison of generated and ground truth trajectories (Case 4)}
  \label{fig:3d_trajectory20}
\end{figure}

\begin{figure}[htb]
  \centering
  \includegraphics[width=1.0\linewidth]{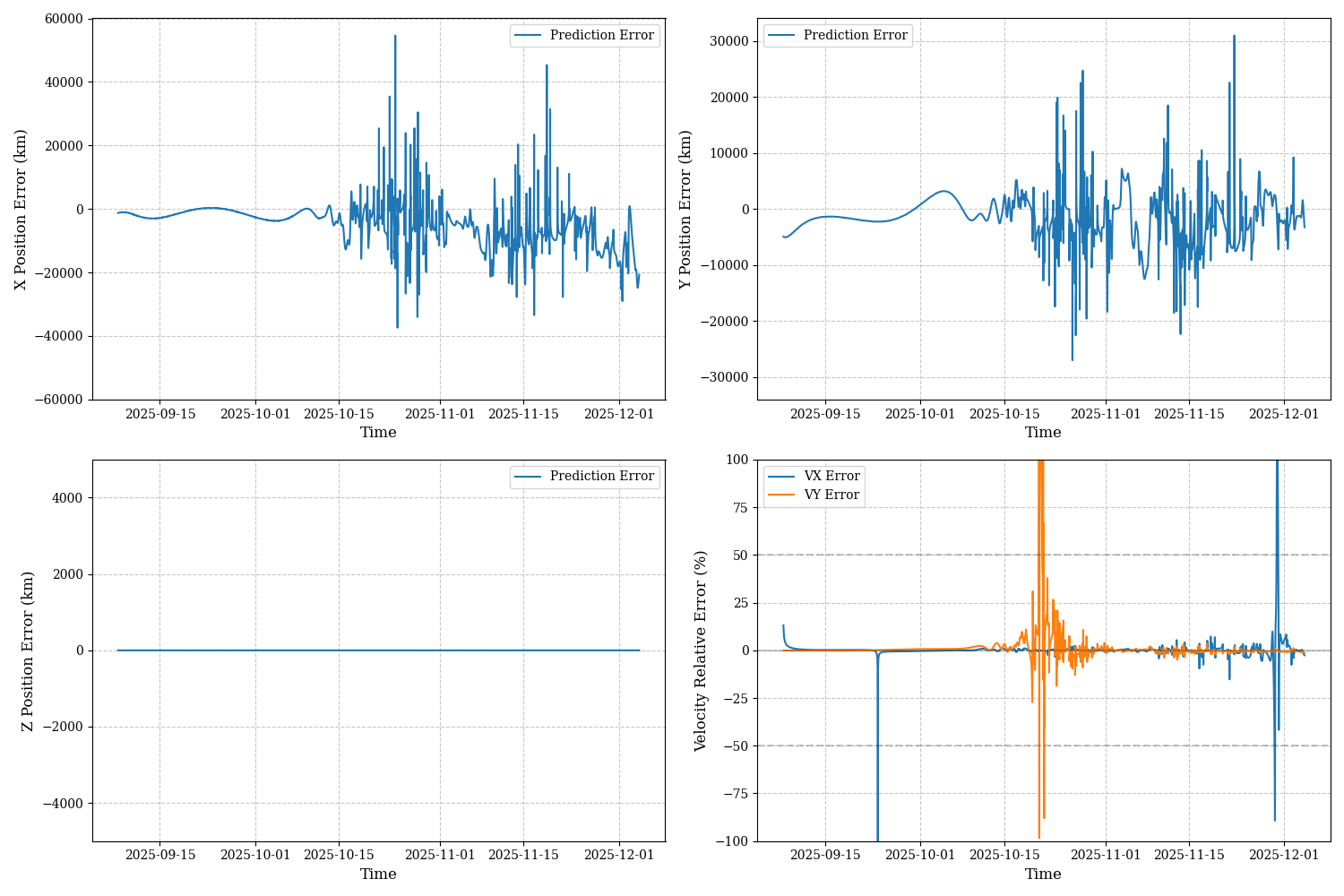}
  \caption{Error time evolution of state variables (Case 4)}
  \label{fig:state_evolution20}
\end{figure}

\begin{figure}[htb]
  \centering
  \includegraphics[width=1.0\linewidth]{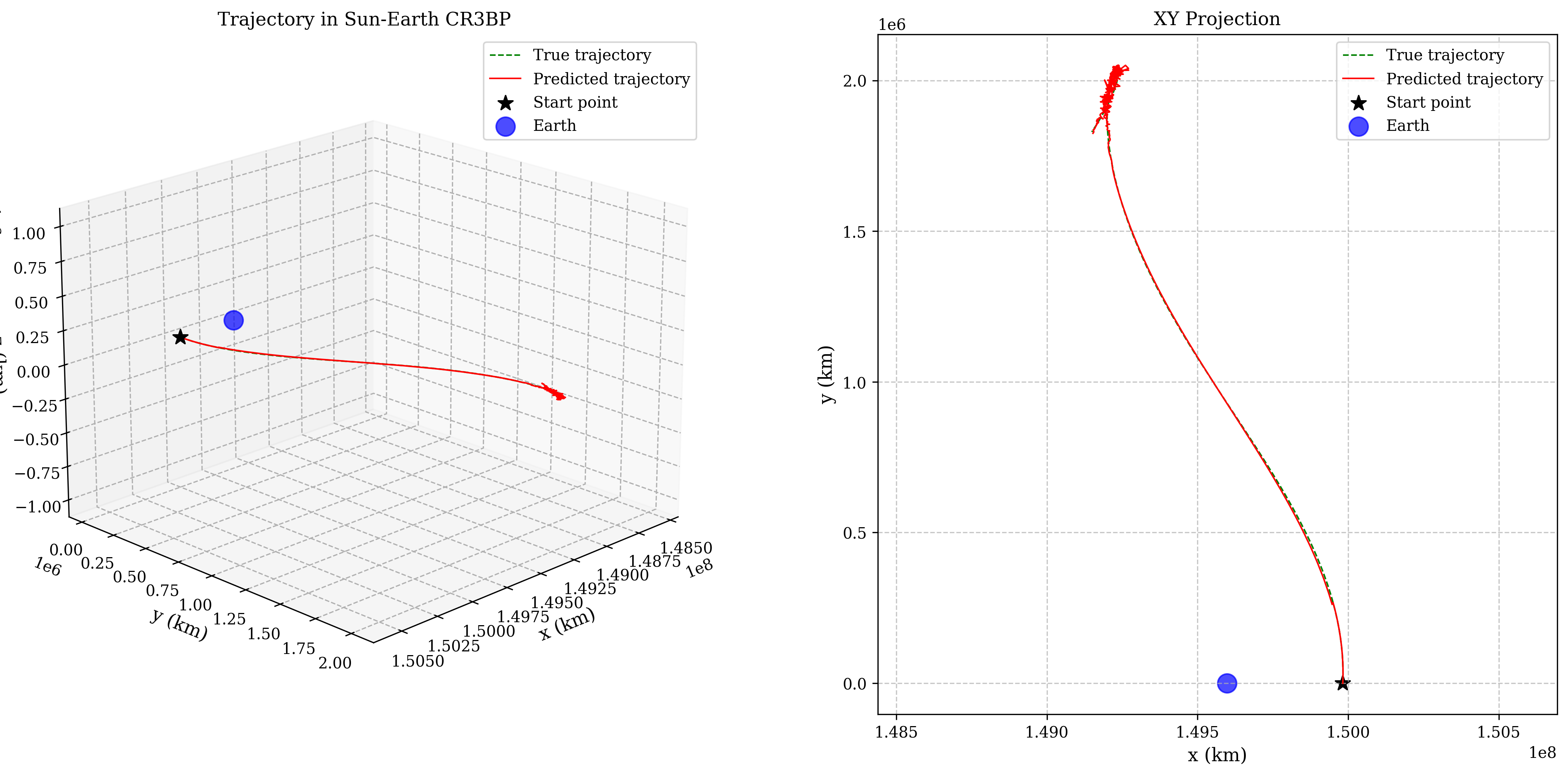}
  \caption{Three-dimensional comparison of generated and ground truth trajectories (Case 5)}
  \label{fig:3d_trajectory22}
\end{figure}

\begin{figure}[htb]
  \centering
  \includegraphics[width=1.0\linewidth]{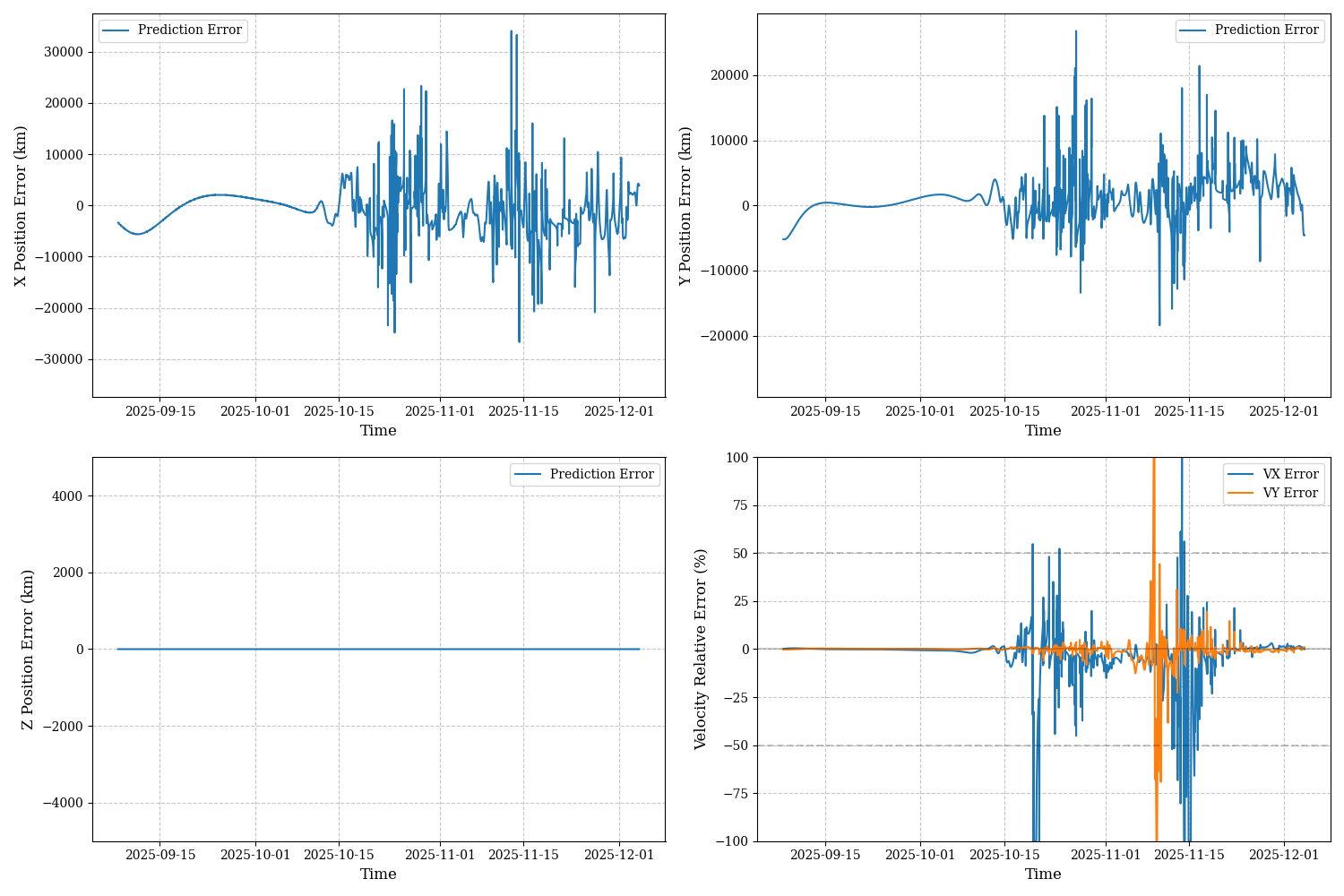}
  \caption{Error time evolution of state variables (Case 5)}
  \label{fig:state_evolution22}
\end{figure}

\begin{figure}[htb]
  \centering
  \includegraphics[width=1.0\linewidth]{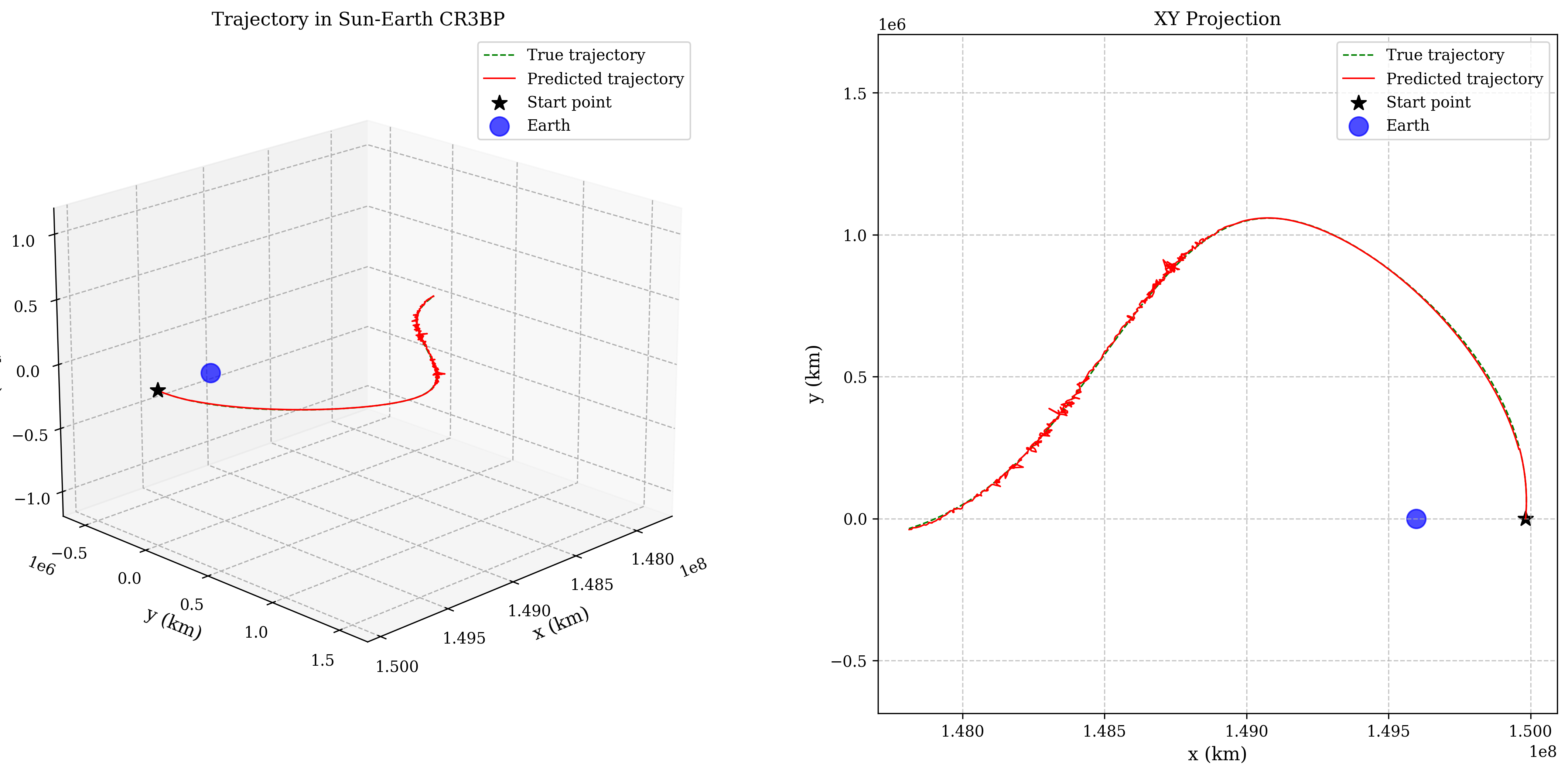}
  \caption{Three-dimensional comparison of generated and ground truth trajectories (Case 6)}
  \label{fig:3d_trajectory32}
\end{figure}

\begin{figure}[htb]
  \centering
  \includegraphics[width=1.0\linewidth]{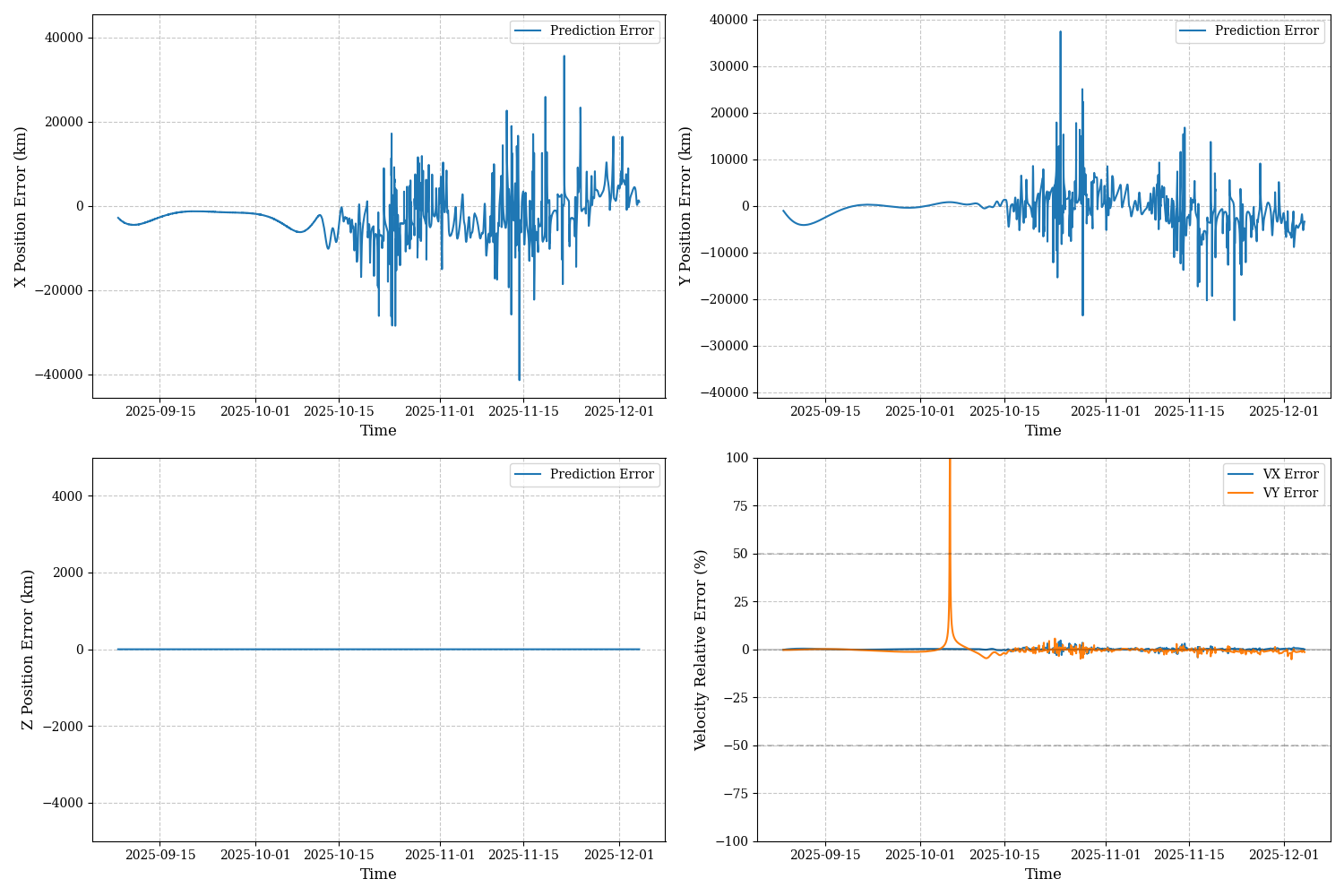}
  \caption{Error time evolution of state variables (Case 6)}
  \label{fig:state_evolution32}
\end{figure}

\begin{figure}[htb]
  \centering
  \includegraphics[width=1.0\linewidth]{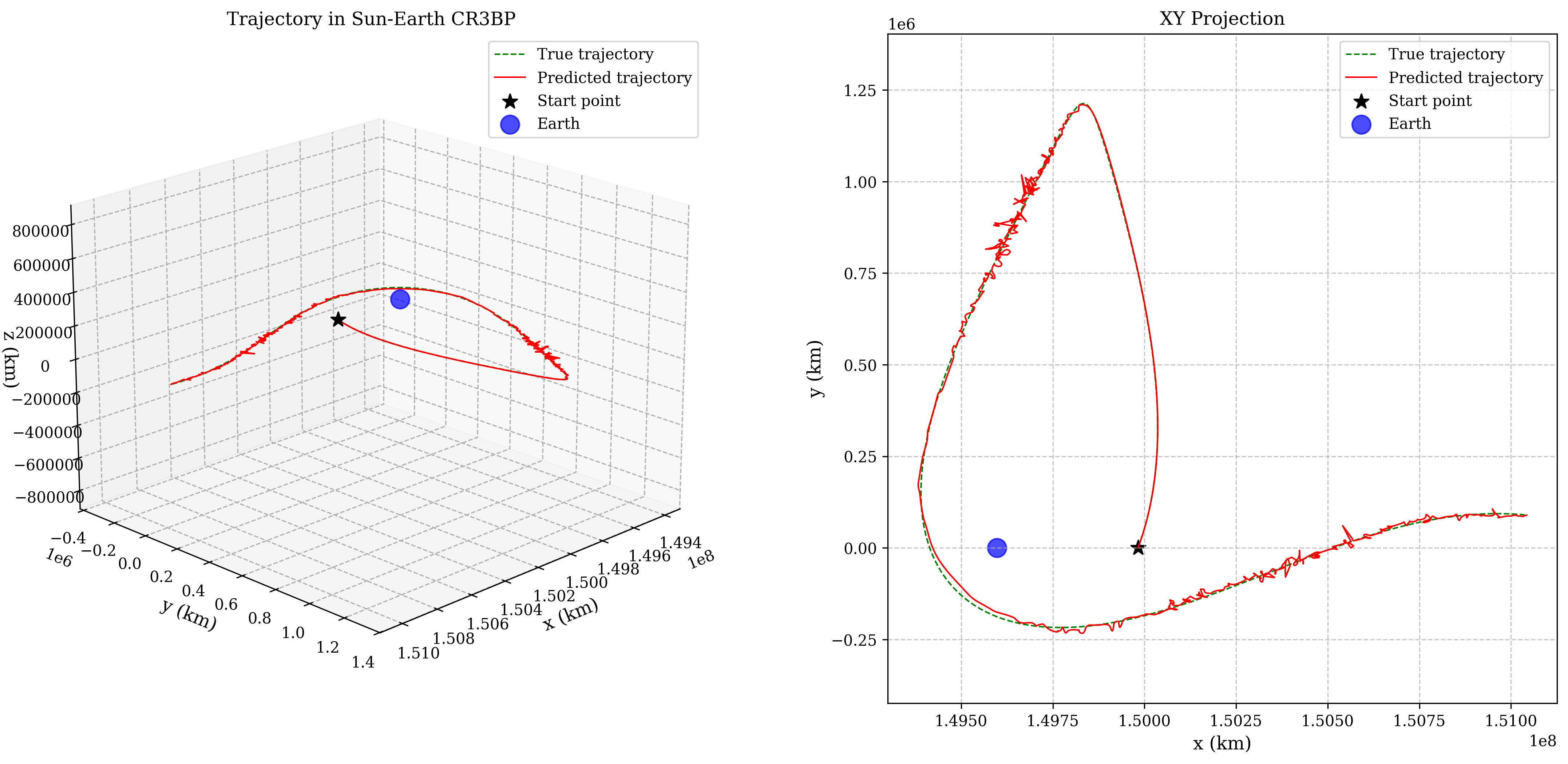}
  \caption{Three-dimensional comparison of generated and ground truth trajectories (Case 7)}
  \label{fig:3d_trajectory34}
\end{figure}

\begin{figure}[htb]
  \centering
  \includegraphics[width=1.0\linewidth]{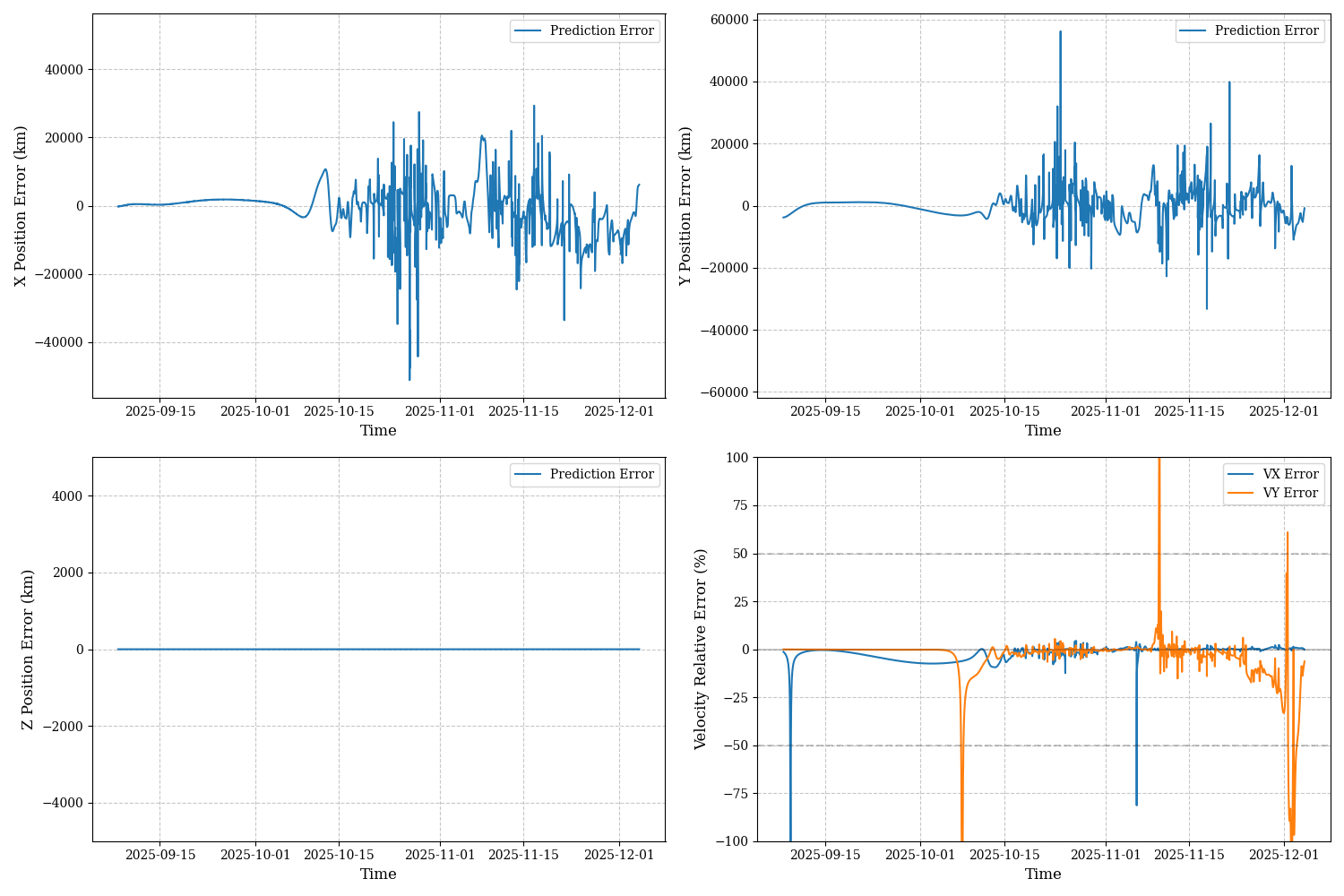}
  \caption{Error time evolution of state variables (Case 7)}
  \label{fig:state_evolution34}
\end{figure}

\begin{figure}[htb]
  \centering
  \includegraphics[width=1.0\linewidth]{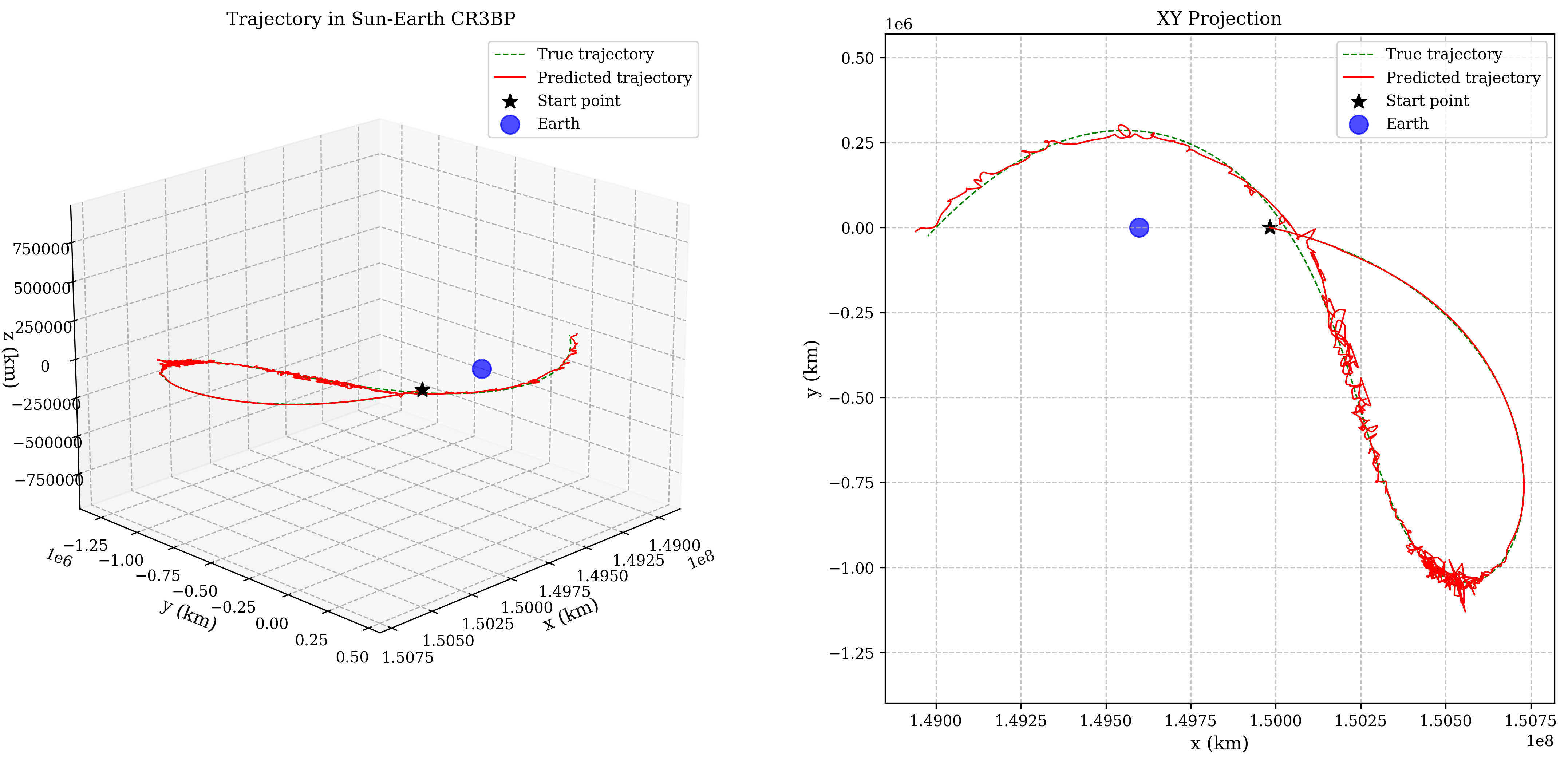}
  \caption{Three-dimensional comparison of generated and ground truth trajectories (Case 8)}
  \label{fig:3d_trajectory40}
\end{figure}

\begin{figure}[htb]
  \centering
  \includegraphics[width=1.0\linewidth]{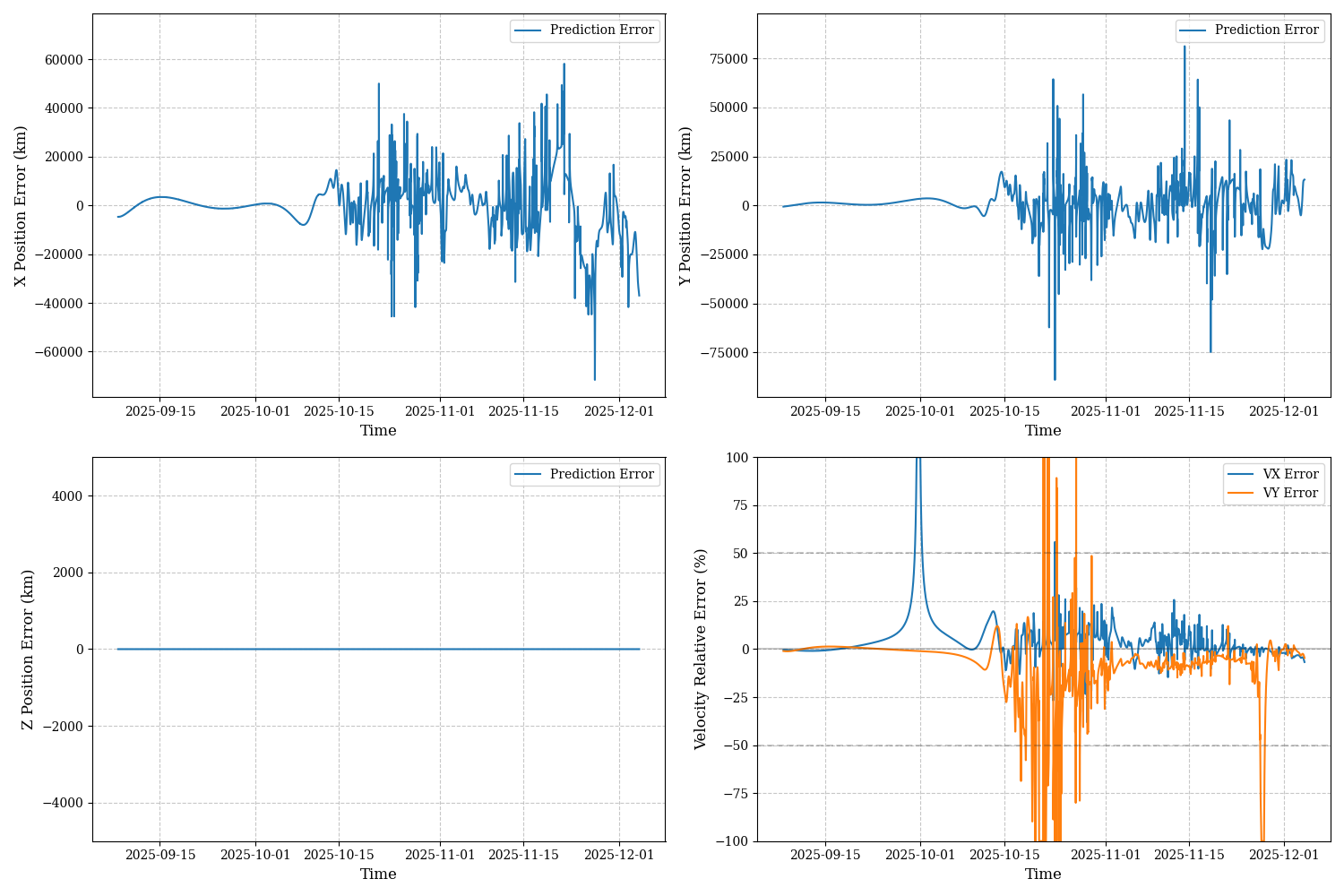}
  \caption{Error time evolution of state variables (Case 8)}
  \label{fig:state_evolution40}
\end{figure}

\begin{figure}[htb]
  \centering
  \includegraphics[width=1.0\linewidth]{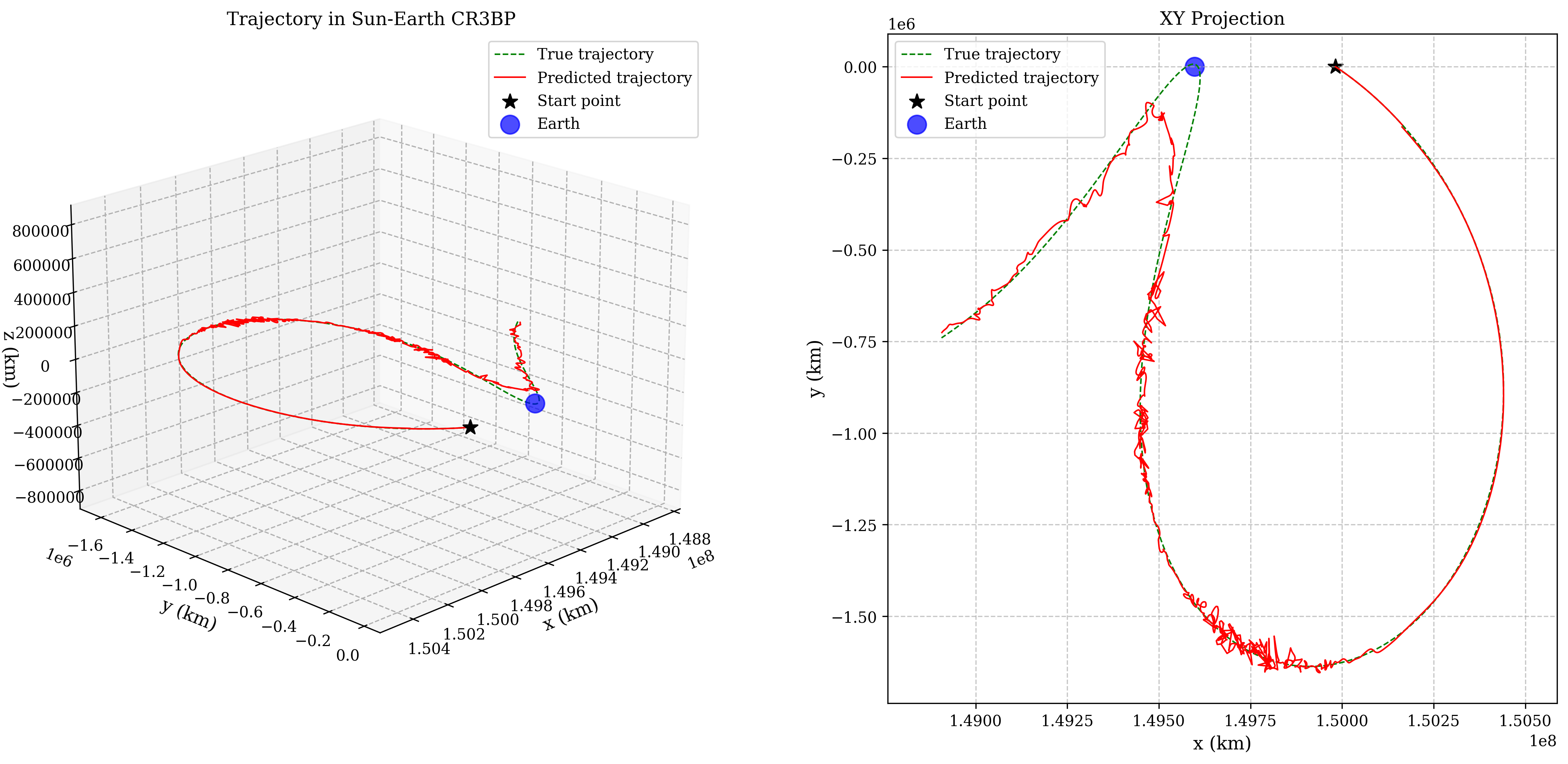}
  \caption{Three-dimensional comparison of generated and ground truth trajectories (Case 9)}
  \label{fig:3d_trajectory50}
\end{figure}

\begin{figure}[htb]
  \centering
  \includegraphics[width=1.0\linewidth]{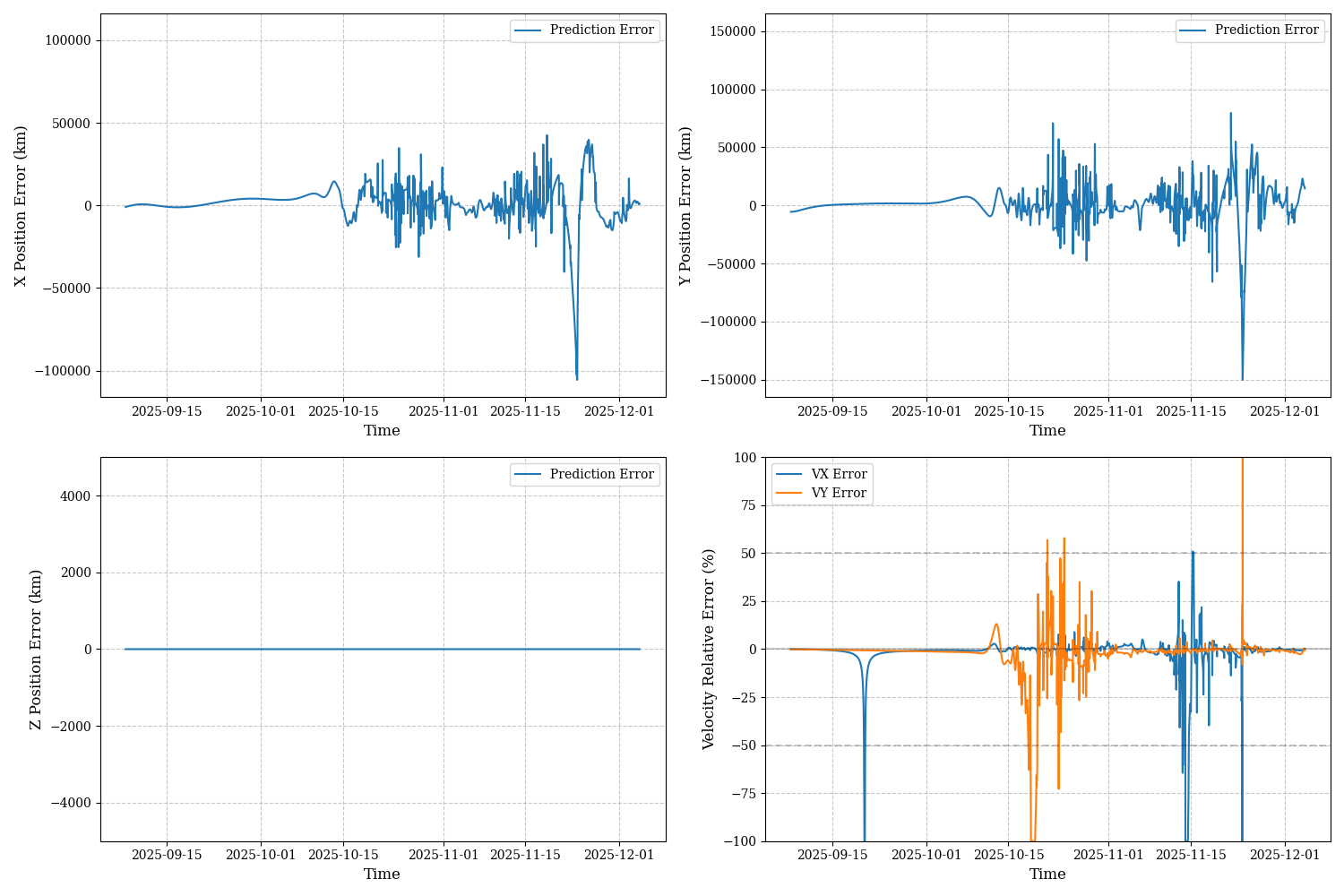}
  \caption{Error time evolution of state variables (Case 9)}
  \label{fig:state_evolution50}
\end{figure}

\begin{figure}[htb]
  \centering
  \includegraphics[width=1.0\linewidth]{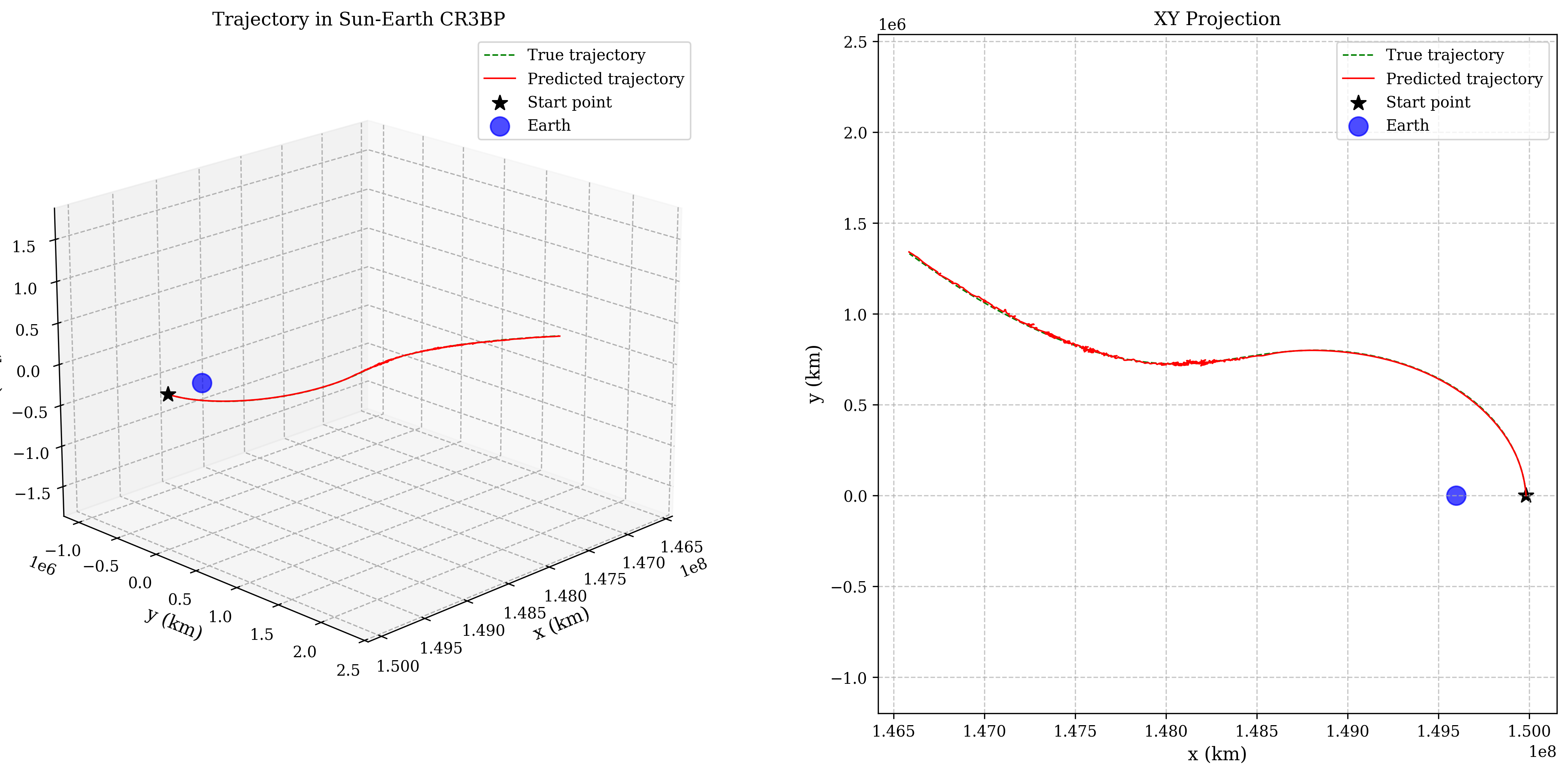}
  \caption{Three-dimensional comparison of generated and ground truth trajectories (Case 10)}
  \label{fig:3d_trajectory58}
\end{figure}

\begin{figure}[htb]
  \centering
  \includegraphics[width=1.0\linewidth]{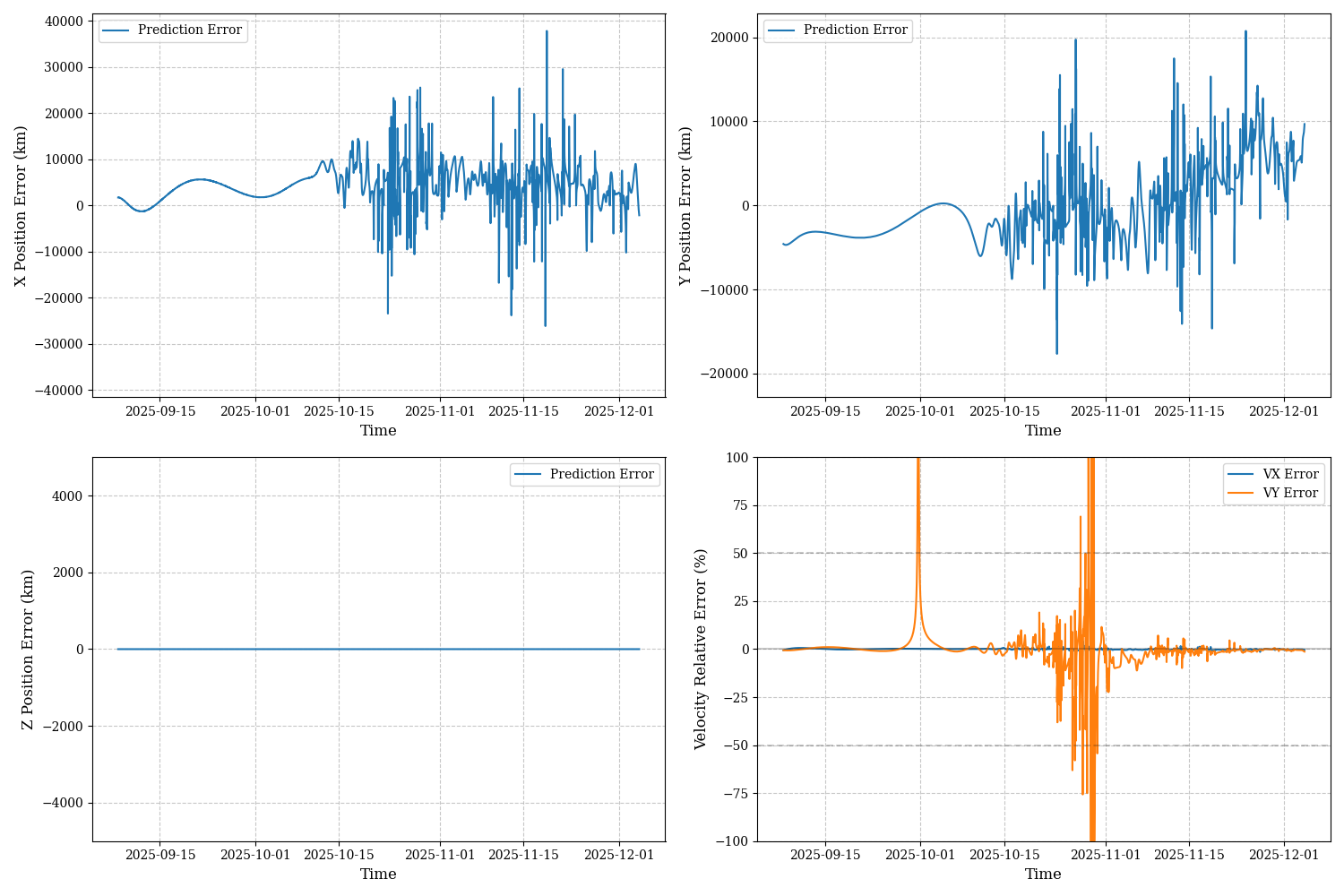}
  \caption{Error time evolution of state variables (Case 10)}
  \label{fig:state_evolution58}
\end{figure}

\begin{figure}[htb]
  \centering
  \includegraphics[width=1.0\linewidth]{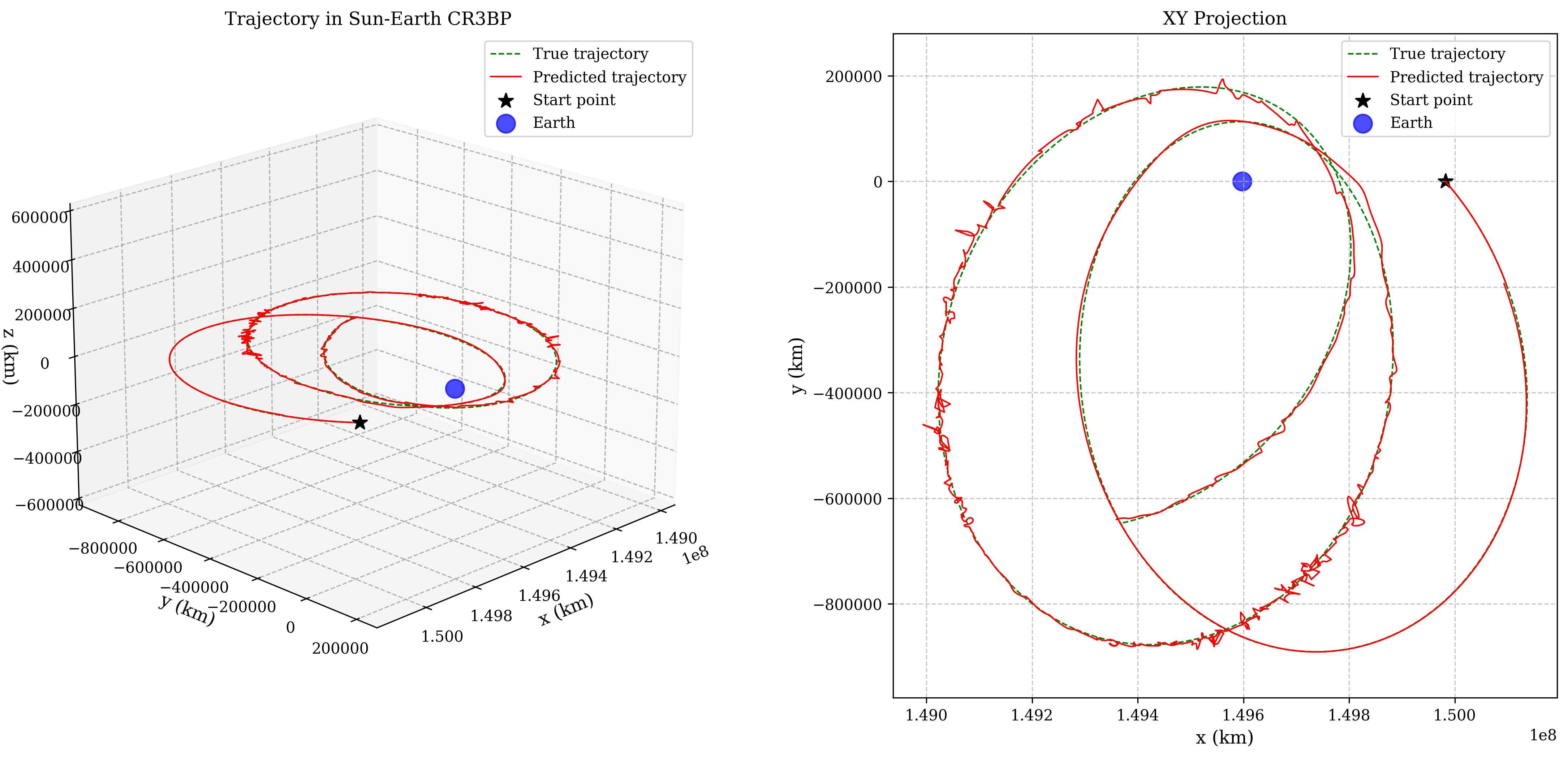}
  \caption{Three-dimensional comparison of generated and ground truth trajectories (Case 11)}
  \label{fig:3d_trajectory66}
\end{figure}

\begin{figure}[htb]
  \centering
  \includegraphics[width=1.0\linewidth]{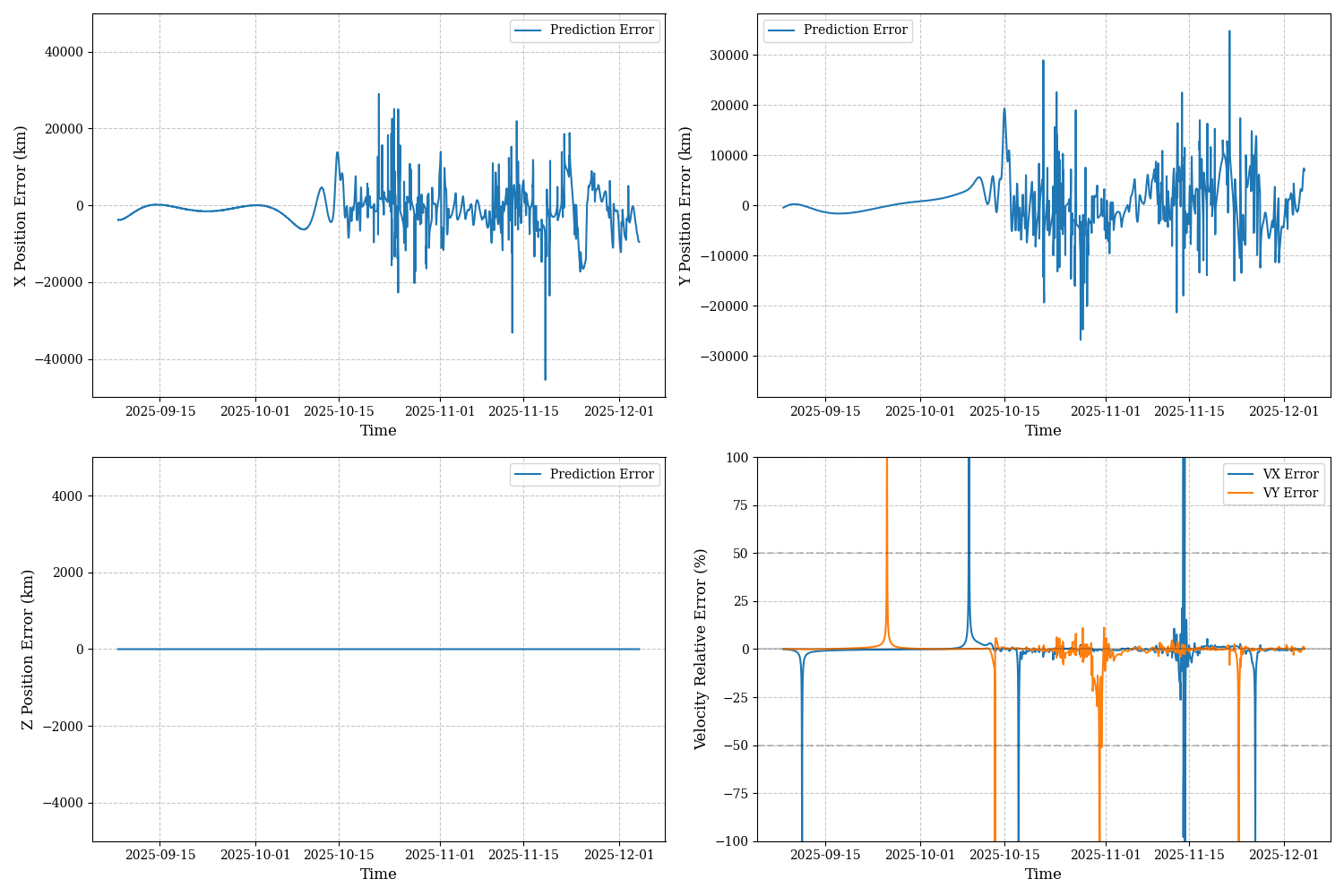}
  \caption{Error time evolution of state variables (Case 11)}
  \label{fig:state_evolution66}
\end{figure}

\begin{figure}[htb]
  \centering
  \includegraphics[width=1.0\linewidth]{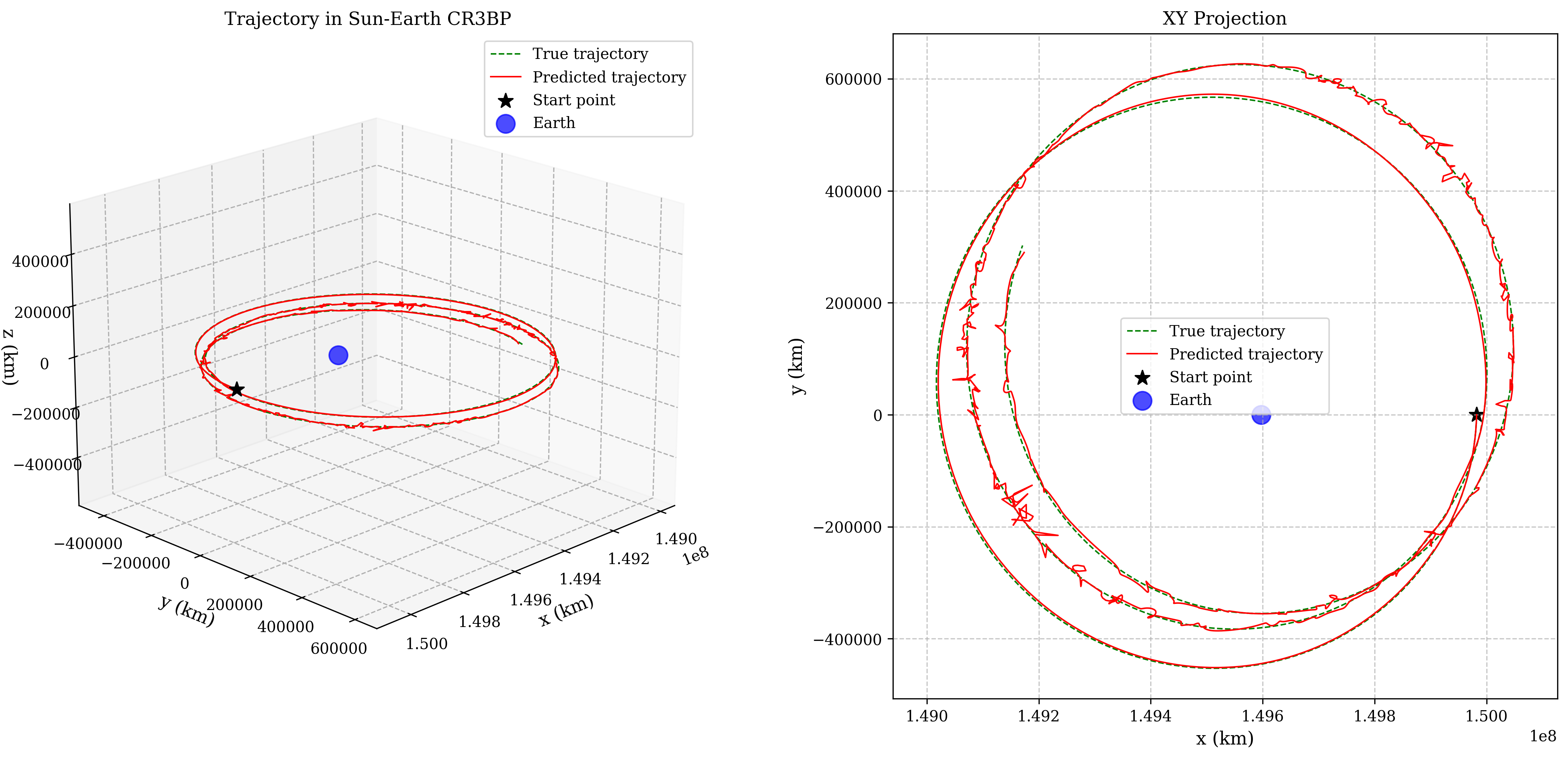}
  \caption{Three-dimensional comparison of generated and ground truth trajectories (Case 12)}
  \label{fig:3d_trajectory75}
\end{figure}

\begin{figure}[htb]
  \centering
  \includegraphics[width=1.0\linewidth]{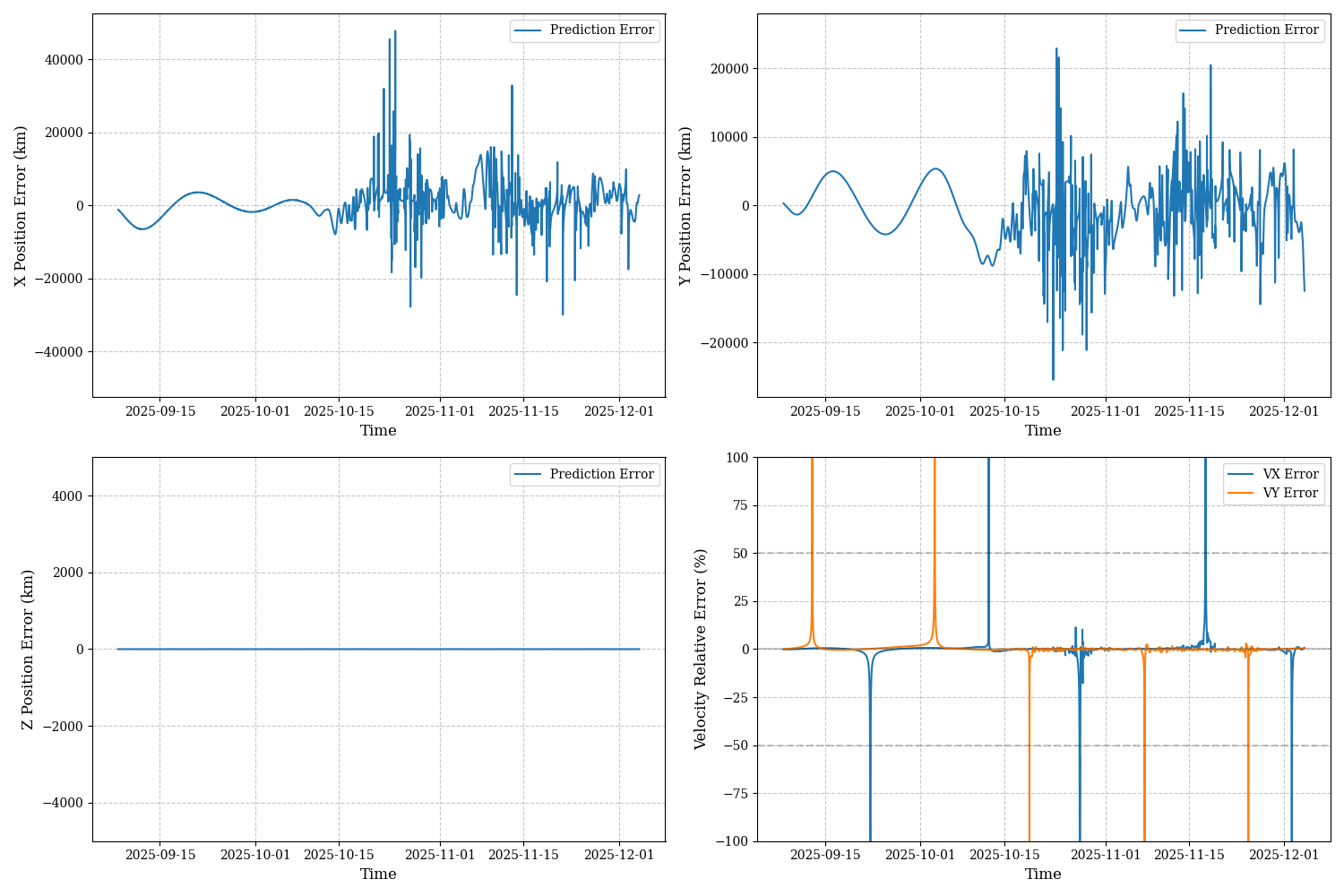}
  \caption{Error time evolution of state variables (Case 12)}
  \label{fig:state_evolution75}
\end{figure}

\begin{figure}[htb]
  \centering
  \includegraphics[width=1.0\linewidth]{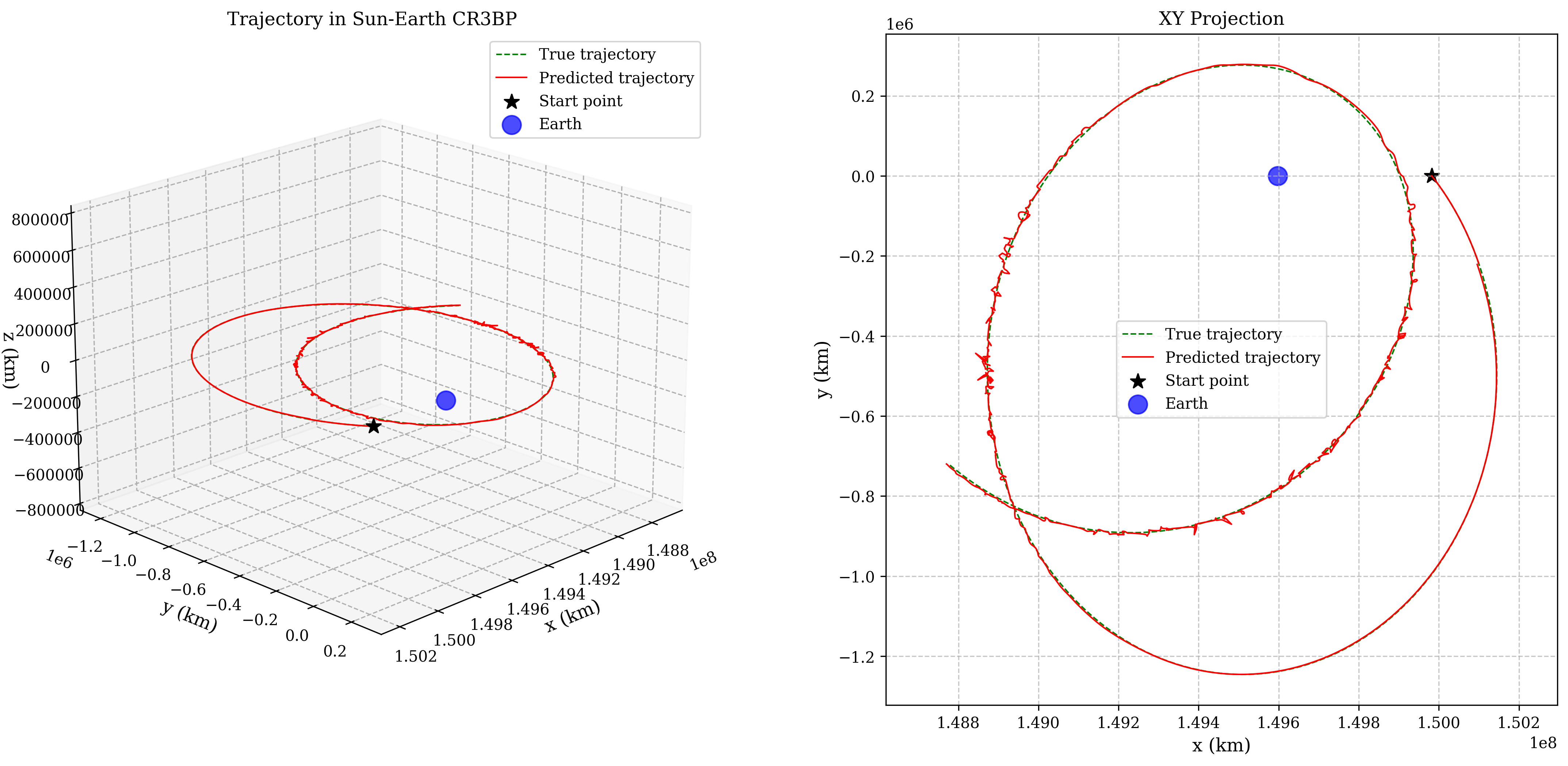}
  \caption{Three-dimensional comparison of generated and ground truth trajectories (Case 13)}
  \label{fig:3d_trajectory99}
\end{figure}

\begin{figure}[htb]
  \centering
  \includegraphics[width=1.0\linewidth]{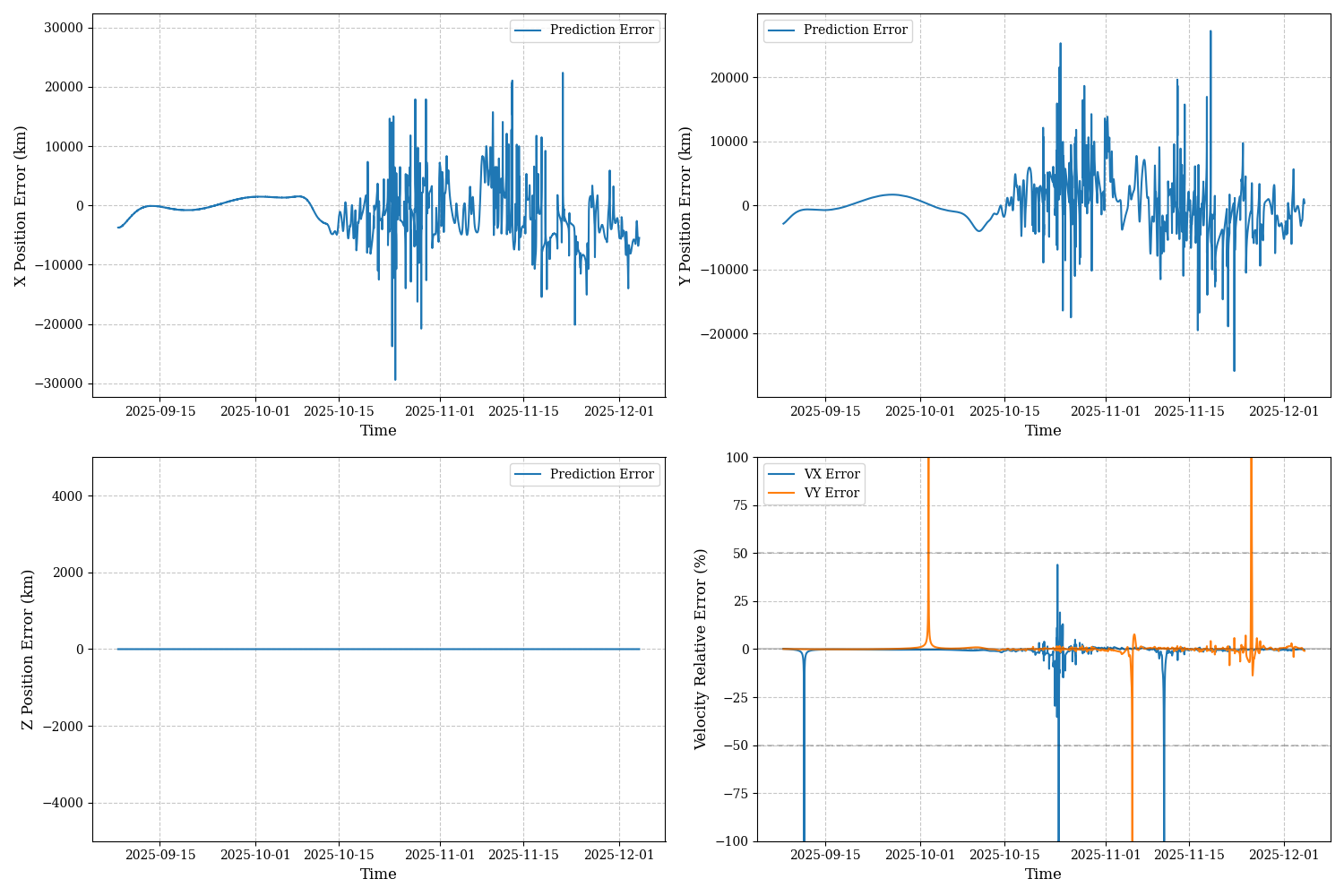}
  \caption{Error time evolution of state variables (Case 13)}
  \label{fig:state_evolution99}
\end{figure}

This section shows detail results of 13 characteristic trajectories, as illustrated in Figure~\ref{fig:3d_trajectory1} to Figure~\ref{fig:state_evolution99}. Among the 100 trajectories generated with different prefix conditions, some results could not be included due to space limitations.

\end{document}